\documentclass[conference]{IEEEtran}
\IEEEoverridecommandlockouts
\usepackage{cite}
\usepackage{amsmath,amssymb,amsfonts}
\usepackage{pgfplots}
\usepackage{amsmath,amssymb}
\usepackage{tikz}
\usetikzlibrary{arrows.meta,positioning,fit,shapes.misc,calc}
\usepackage{cite}      % compresses [1]-[3] etc.
\usepackage{url}       % for URLs in bib entries
\usepackage{pifont}

\pgfplotsset{compat=1.18} % or your installed version % or your installed version
\usepackage{graphicx}
\usepackage{medmath}
\usepackage{adjustbox}
\usepackage{textcomp}
\usepackage{xcolor}
\usepackage{amsmath, amssymb, amsfonts, bm}
\usepackage{graphicx}
\usepackage{booktabs}
\usepackage{multirow}
\usepackage{hyperref}
\usepackage{enumitem}
\usepackage{mathtools}
\usepackage{algorithm}
\usepackage{algpseudocode}
\usepackage{siunitx}
\usepackage{xcolor}

\usepackage{placeins}    % \FloatBarrier to flush floats
\usepackage{dblfloatfix} % improves two-column float placement
\graphicspath{{./}{./figs/}} % adjust if your images are in a folder
\DeclareGraphicsExtensions{.pdf,.png,.jpg}
\usepackage{microtype}
\usepackage{subcaption}
\usepackage{balance}
\def\BibTeX{{\rm B\kern-.05em{\sc i\kern-.025em b}\kern-.08em
    T\kern-.1667em\lower.7ex\hbox{E}\kern-.125emX}}
\newcommand{\medmath}[1]{\ensuremath{#1}}

\newcommand{\ExCIR}{\textsc{ExCIR}}

\begin{document}

\title{ Correlation-Aware Feature Attribution Based Explainable AI}
% ~\\
% \and
% \IEEEauthorblockN{2\textsuperscript{nd} Given Name Surname*}
% *Corresponding author
% ~\\
% \and
% ~\\
% \and
% \and

\author{
\IEEEauthorblockN{
    Poushali Sengupta*,
    Yan Zhang,
    Frank Eliassen**,
    Sabita Maharjan***
}
\IEEEauthorblockA{
    Department of Informatics, University of Oslo, Norway\thanks{This project is supported by dScience: the Centre for Computational and Data Science, University of Oslo.}
 \\
    Emails: poushali@ifi.uio.no, 
           yanzhang@ieee.org,
            frank@ifi.uio.no,
            sabita@ifi.uio.no
\\ ORCID: *0000-0002-6974-5794, **0000-0002-7788-4137, ***0000-0002-4616-8488
}
}

\maketitle

\begin{abstract}

Explainable AI (XAI) is increasingly essential as modern models become more complex and high-stakes applications demand transparency, trust, and regulatory compliance. Existing global attribution methods often incur high computational costs, lack stability under correlated inputs, and fail to scale efficiently to large or heterogeneous datasets. We address these gaps with \emph{ExCIR} (Explainability through Correlation Impact Ratio), a correlation-aware attribution score equipped with a lightweight transfer protocol that reproduces full-model rankings using only a fraction of the data. ExCIR quantifies sign-aligned co-movement between features and model outputs after \emph{robust centering} (subtracting a robust location estimate, e.g., median or mid-mean, from features and outputs). We further introduce \textsc{BlockCIR}, a \emph{groupwise} extension of ExCIR that scores \emph{sets} of correlated features as a single unit. By aggregating the same signed-co-movement numerators and magnitudes over predefined or data-driven groups, \textsc{BlockCIR} mitigates double-counting in collinear clusters (e.g., synonyms or duplicated sensors) and yields smoother, more stable rankings when strong dependencies are present. Across diverse text, tabular, signal, and image datasets, ExCIR shows trustworthy agreement with established global baselines and the full model, delivers consistent top-$k$ rankings across settings, and reduces runtime via lightweight evaluation on a subset of rows. Overall, ExCIR provides \emph{computationally efficient}, \emph{consistent}, and \emph{scalable} explainability for real-world deployment.

\end{abstract}


% ================================
% ================================

% ================================
\section{Introduction}

As Machine Learning systems increasingly operate in safety-critical, regulated, and user-facing settings, stakeholders demand transparent and auditable reasoning to support trust, accountability, debugging, drift monitoring, and decision support. Consequently, trustworthy and computationally efficient explanations are essential, and post-hoc feature-attribution methods are widely used to interpret black-box models across text, tabular, image, and signal domains.
Despite their popularity, many such pipelines remain (i) \emph{computationally heavy}, often requiring hundreds to thousands of re-evaluations of the model;  
(ii) \emph{brittle under cross-feature dependence}, which can cause instability and double-counting in correlated inputs~\cite{michalak2010correlation}; and  
(iii) \emph{difficult to deploy under privacy or latency constraints}, particularly when explanations must be generated  near real-time and/or on resource limited devices.  
In practical analytics and decision-support settings, such as clinical diagnostics, IoT-driven monitoring, or autonomous driving, users require explanations that are  
\emph{low in computational complexity}, \emph{consistent}~\cite{alvarezmelis2018robustness,yeh2019infidelity}, and \emph{scalable} without repeatedly querying the model.  
We target global feature ranking that is calibration-invariant and efficient at large datasets without model re-evaluation, enabling analysts to obtain explanations that are \emph{low complexity}, \emph{consistent}, and \emph{scalable} even in resource-constrained environments.

Popular perturbation- or sampling-based methods such as LIME~\cite{ribeiro2016lime}, SHAP~\cite{lundberg2017shap}, permutation feature importance (PFI)~\cite{breiman2001randomforest}, and PDP-variance~\cite{friedman2001pdp} trade faithfulness for compute budgets and rely on background-distribution assumptions that may be invalid under covariate shift.  
Gradient-based saliency maps~\cite{simonyan2013saliency} and Integrated Gradients~\cite{sundararajan2017integratedgradients} are efficient but can be sensitive to scaling, saturation, or baseline choice, and they do not directly address cross-feature dependence.  
Global dependence measures such as the Hilbert–Schmidt Independence Criterion (HSIC)~\cite{gretton2005hsic}, mutual information with labels or predictions~\cite{ross2014mi}, or model-intrinsic statistics such as TreeGain~\cite{chen2016xgboost} and surrogate-LR coefficients~\cite{molnar2020interpretable} offer valuable global views, but they may be expensive to compute repeatedly and can double-count signal in highly collinear feature groups.  
A practical solution should therefore (a) provide \emph{interpretable feature scores} transferring across model calibrations, (b) \emph{handle correlated groups} without redundant attribution, and (c)  \emph{lower} computational cost.
To address the above limitations, we propose ExCIR: Explainability through Correlation Impact Ratio, a correlation-aware attribution score that quantifies the \emph{sign-aligned co-movement} between each feature and the model output after a robust mid-mean centering step~\cite{huber1981robust,hampel1971robust}.  
ExCIR makes a single linear scan over the kept rows, runtime is proportional to the number of observations $n$ (with a per-row cost across $d$ features), and naturally extends to correlated feature groups through \emph{\textsc{BlockCIR}}.  
To reduce cost at scale, we introduce a \emph{lightweight transfer} protocol that reproduces the full-model feature-ranking geometry while using only a fraction (20–40\%) of the data.  
This controlled subsampling achieves multi-fold speed-ups with minimal loss of agreement, quantified \emph{non-perturbatively} via top-$k$ Jaccard overlap~\cite{jaccard1901etude}, rank correlations (Spearman/Kendall)~\cite{spearman1904,kendall1938}, shape alignment (orthogonal Procrustes residual)~\cite{schonemann1966procrustes}, and distributional similarity (symmetric KL)~\cite{jeffreys1946}.
The method is practical for iterative workflows and deployment on edge hardware, as well as streaming/online settings: it requires only streaming-friendly quantile estimates that can be updated in one pass with small memory footprints (e.g., Greenwald–Khanna and t-digest)~\cite{greenwald2001quantiles}. Our main contributions are:
\begin{itemize}[leftmargin=*,itemsep=2pt,topsep=2pt]
  \item \emph{\textbf{Systematic formulation.}} A unified \emph{Correlation Impact Ratio} [$\mathrm{CIR}(G)$] for sets of features (single-pass; translation/scale invariant; sign-symmetric; monotone response).

  \item \emph{\textbf{Correlation-aware attribution.}} A setwise scoring method,
\emph{\textsc{BlockCIR}} that aggregates related features to avoid double counting.
  \item \emph{\textbf{Consistancy \& efficiency.}} A lightweight transfer protocol that shows strong agreement (rank overlap, shape alignment, distributional match) at substantially lower cost.
\end{itemize}

\paragraph{Paper roadmap.}
\autoref{sec:method} defines the setup and notation, introduces the co-movement primitives, and formalizes \(\mathrm{CIR}(G)\) with core properties, a one-pass algorithm, the \textsc{BlockCIR} and class-conditioned variants, and the lightweight protocol (including dataset/model setup).
\autoref{sec:results} reports evaluation across diverse datasets, focusing on agreement, runtime and cost–agreement trade-offs, feature-importance distributions, and concise ablations. We omit perturbation-based metrics (e.g., AOPC) and focus on agreement, shape alignment, and distributional match; ethics and reproducibility appear last.

\section{Method: \ExCIR}
\label{sec:method}
\subsection{Notation}
\label{sec:notation}

Table~\ref{tab:notation} summarizes the symbols used throughout the paper.
Big-$\Theta(\cdot)$ indicates asymptotic order: $T=\Theta(n\,d)$ grows
linearly with both the number of samples $n$ and features $d$ up to constants.

\begin{table}[h]
\centering
\caption{Notation used in the paper.}
\label{tab:notation}
\renewcommand{\arraystretch}{1.05}
\setlength{\tabcolsep}{4pt}
\begin{adjustbox}{width = \linewidth}
\begin{tabular}{ll}
\toprule
\textbf{Symbol} & \textbf{Description} \\ \midrule
$n$ & Number of samples (rows) in the dataset \\
$d$ & Number of input features (columns) \\
$\bm{x}_i \in \mathbb{R}^{d}$ & Feature vector of sample $i$ \\
$x_{ij}$ & $j$-th feature value of sample $i$ \\
$\hat{\bm{y}} \in \mathbb{R}^{n}$ & Model predictions on all $n$ samples \\
$\tilde{x}_{ij},\,\tilde{y}_{i}$ & Robustly centered values (mid-mean) \\
$m(\cdot)$ & Mid-mean operator (trimmed mean of central 50\%) \\
$G \subseteq \{1,\dots,d\}$ & Correlated feature group; $\mathcal{G}$ all groups \\
$p_{ij}=\tilde{x}_{ij}\tilde{y}_{i}$ & Signed co-movement term for feature $j$ \\
$N_j=\sum_{i}p_{ij}$,\; $D_j=\sum_{i}|p_{ij}|$ & Accumulators for feature $j$ \\
$\mathrm{CIR}_j=\tfrac12(1+N_j/D_j)$ & ExCIR score for feature $j$ ($\in[0,1]$) \\
$\mathrm{\textsc{BlockCIR}}_G$ & Groupwise CIR score for block $G$ \\
$f\in(0,1]$ & Kept-row fraction in lightweight transfer \\
$T_{\text{center}}$ & Robust-centering cost:
$\Theta(d\,n\log n)$ (sorting) or $\Theta(d\,n)$ (streaming) \\
$T_{\text{one-pass}}$ & One-pass accumulation cost $\Theta(n\,d)$ \\
Memory & $\Theta(d)$ for features ($+\Theta(|\mathcal{G}|)$ for groups) \\
\bottomrule
\end{tabular}
\end{adjustbox}
\end{table}

\subsection{Setup and goal}
We consider a dataset with $n$ samples and $d$ features. Let $X\in\mathbb{R}^{n\times d}$ be the feature matrix with column $j$ denoted $\bm{x}_j\in\mathbb{R}^n$, and let $f:\mathbb{R}^d\!\to\!\mathbb{R}$ be a trained model with outputs $\bm{y}=f(X)\in\mathbb{R}^n$ (for multi-class, use class scores $f_c(X)$; see \autoref{sec:class}). 
Our goal is to assign each feature a score that captures \emph{how consistently} it moves in the same direction as the model output after removing simple location effects via robust centering.
We design a score that (i) is immediately interpretable as aligned co-movement, (ii) is translation- and positive scale-invariant in both arguments (insensitive to re-centering/re-scaling), (iii) is computable in a single pass without re-evaluating the model, and (iv) extends naturally to blocks of correlated features. \autoref{fig:pipeline} illustrates our methodolgy pipeline. To situate ExCIR among common explainers, Table~\ref{tab:sota_excir} summarizes typical trade-offs in computational cost, invariance, and correlation handling. Let $[n] = \{1,\dots,n\}$ and $[d]=\{1,\dots,d\}$. 
Bold lower-case ($\bm{a}$) denote vectors and bold upper-case ($\bm{A}$) matrices; 
$\bm{a}\odot\bm{b}$ is the elementwise product; 
$\|\bm{a}\|_1=\sum_{i\in[n]}|a_i|$; 
$\mathrm{sign}(0)=0$; 
$\mathbb{I}\{\cdot\}$ is the indicator.
A \emph{feature set} is any nonempty $G\subseteq[d]$ with cardinality $|G|$; a family of sets is $\mathcal{G}\subseteq\mathcal{P}([d])$ (not necessarily a partition unless stated). 
For $j\in[d]$, $\bm{x}_j\in\mathbb{R}^n$ is column $j$ of $X$; for $G\subseteq[d]$, $X_G=(x_{ij})_{i\in[n],\,j\in G}$.
We center vectors by the mid-mean $m(\bm{a})=\tfrac{1}{2}\!\left(Q_{0.25}(\bm{a})+Q_{0.75}(\bm{a})\right)$ and write 
$\tilde{\bm{x}}_j=\bm{x}_j-m(\bm{x}_j)\bm{1}$, $\tilde{\bm{y}}=\bm{y}-m(\bm{y})\bm{1}$.

% ---------- small schematic ----------
% Preamble
\usetikzlibrary{positioning,calc,arrows.meta}

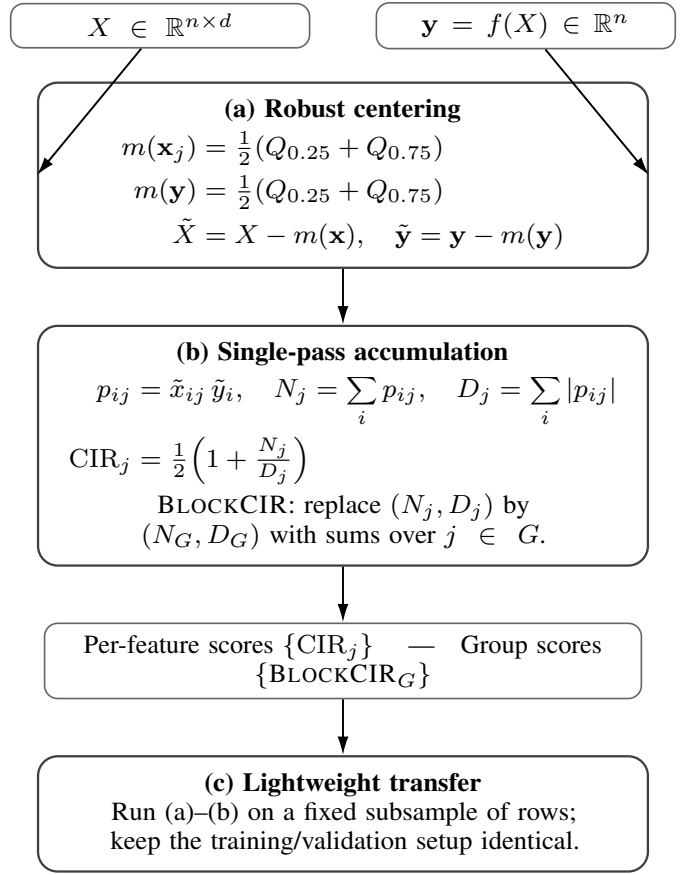
\begin{figure}[t]
\centering
\resizebox{\linewidth}{!}{%
\begin{tikzpicture}[
  node distance=5mm and 6mm,
  every node/.style={font=\footnotesize},
  arr/.style={-{Latex[length=2mm,width=1.2mm]}, semithick},
  stage/.style={rounded corners=6pt, draw=black!75, line width=0.7pt, fill=white,
                inner sep=5pt, align=center, text width=6.2cm},
  chip/.style ={rounded corners=4pt, draw=black!60, line width=0.5pt, fill=white,
                inner sep=3pt, align=center, text width=6.2cm}
]

% Inputs
\node[chip, text width=3.0cm] (X) {$X\in\mathbb{R}^{n\times d}$};
\node[chip, text width=3.0cm, right=7mm of X] (y) {$\mathbf{y}=f(X)\in\mathbb{R}^{n}$};

% Midpoint helper (prevents the old “No shape named ($(X” error and stray chars)
\coordinate (midXY) at ($(X)!0.5!(y)$);

% (a) Robust centering
\node[stage, below=6mm of midXY] (center) {\textbf{(a) Robust centering}\\[1mm]
$\begin{aligned}
m(\mathbf{x}_j) &= \tfrac{1}{2}\!\left(Q_{0.25}+Q_{0.75}\right)\\
m(\mathbf{y})  &= \tfrac{1}{2}\!\left(Q_{0.25}+Q_{0.75}\right)\\
\tilde{X}      &= X - m(\mathbf{x}),\quad
\tilde{\mathbf{y}} = \mathbf{y} - m(\mathbf{y})
\end{aligned}$};
\draw[arr] (X) -- (center.west);
\draw[arr] (y) -- (center.east);

% (b) Single-pass accumulation
\node[stage, below=6mm of center] (onepass) {\textbf{(b) Single-pass accumulation}\\[1mm]
$\begin{aligned}
p_{ij} &= \tilde{x}_{ij}\,\tilde{y}_i, \quad
N_j=\sum_i p_{ij},\quad D_j=\sum_i |p_{ij}|\\
\mathrm{CIR}_j &= \tfrac{1}{2}\!\left(1+\tfrac{N_j}{D_j}\right)
\end{aligned}$\\[1mm]
\textsc{\textsc{BlockCIR}}: replace $(N_j,D_j)$ by $(N_G,D_G)$ with sums over $j\in G$.};
\draw[arr] (center) -- (onepass);

% Outputs
\node[chip, below=6mm of onepass] (outs)
{\begin{minipage}{6.2cm}\centering
Per-feature scores $\{\mathrm{CIR}_j\}$ \quad—\quad Group scores $\{\mathrm{\textsc{BlockCIR}}_G\}$
\end{minipage}};
\draw[arr] (onepass) -- (outs);

% (c) Lightweight transfer (no numbers)
\node[stage, below=6mm of outs] (light)
{\textbf{(c) Lightweight transfer}\\
Run (a)–(b) on a fixed subsample of rows; keep the training/validation setup identical.};
\draw[arr] (outs) -- (light);

\end{tikzpicture}%
}
\caption{\textbf{ExCIR workflow} (a) Robust centering via the mid-mean;
(b) single-pass accumulation for per-feature CIR and groupwise \textsc{BlockCIR}; (c) optional lightweight transfer by subsampling rows under the same training/validation setup. For streams, quantiles can be maintained with Greenwald–Khanna or t-digest.}
\label{fig:pipeline}
\end{figure}

\begin{table}[t]
\centering
\scriptsize
\setlength{\tabcolsep}{4pt}
\begin{adjustbox}{width = \linewidth}
\begin{tabular}{lccc}
\toprule
 & \textbf{Cost} & \textbf{Invariance} & \textbf{Correlation handling} \\
\midrule
Perturb./Occlusion & High (many re-evals) & $\times$ & Partial (ad-hoc) \\
Grad./Saliency     & Low--Med             & $\times$ (scale-sensitive) & Weak \\
Shapley-style      & Very high            & \checkmark~(by def.)       & Double-count risk \\
\midrule
\textbf{ExCIR (ours)} & \textbf{Single-pass} & \textbf{shift + pos.-scale} & \textbf{\textsc{BlockCIR} (groups)} \\
\bottomrule
\end{tabular}
\end{adjustbox}
\caption{SOTA families vs.\ ExCIR. ExCIR is single-pass, shift/positive-scale invariant, and group-aware via \textsc{BlockCIR}.}
\label{tab:sota_excir}
\end{table}
See Fig.~\ref{fig:pipeline} for the overall workflow; we now detail each stage in order.\\
\textbf{(1) Inputs:} Start from the feature matrix $X$ and model outputs $\bm{y}=f(X)$ (for class $c$, use $f_c(X)$). Our goal is a global ranking of features (or groups of features) that is computationally efficient and consistent across settings.\\
\textbf{(2) Robust centering:} We recentre each feature and the output so that trivial shifts and scale effects do not dominate the signal. Using a robust midhinge (Tukey’s mid-mean) dampens the impact of outliers and heavy tails, and can be maintained online with compact sketches, making the procedure streaming-friendly~\cite{tukey1977eda}.\\
\textbf{(3) Aligned co-movement:} Intuitively, a feature is important if it tends to move in the same direction as the model’s output across samples. We summarize this “sign-aligned co-movement” in a single pass to obtain a score per feature that is invariant to translation and positive rescaling, supporting apples-to-apples comparisons.\\
\textbf{(4) Groupwise attribution:} Correlated or redundant features (e.g., synonyms in text, collinear sensors) can split credit and obscure the true signal. We therefore aggregate related features first and then assess alignment at the group level (\textsc{BlockCIR}), which reduces double-counting while preserving leaders.\\
\textbf{(5) Lightweight transfer:} When compute or data access is limited, we run the same procedure on a fixed subsample of rows. This retains the global ranking geometry at a fraction of the cost; we verify agreement with standard metrics (e.g., top-$k$ overlap and rank correlations) without resorting to input perturbations.
\par The precise primitives, the unified set functional \(\mathrm{CIR}(G)\), properties, and the one-pass algorithm follow in \autoref{sec:method}. These properties are unavailable in perturbation-based, kernel, or gain-style explainers without substantial re-computation.

\subsection{Robust Centering}
\label{sec:centering}
To dampen outliers and heavy tails, we center every vector by its \emph{mid-mean}:
\[
m(\bm{a}) \;=\; \tfrac{1}{2}\big(Q_{0.25}(\bm{a}) + Q_{0.75}(\bm{a})\big),
\]
where $Q_\alpha(\bm{a})$ is the empirical $\alpha$-quantile of $\bm{a}$.
Define centered features and outputs, for each $j\in[d]$,
\[
\tilde{\bm{x}}_j \;=\; \bm{x}_j - m(\bm{x}_j)\bm{1},
\qquad
\tilde{\bm{y}} \;=\; \bm{y} - m(\bm{y})\bm{1},
\]
with $\bm{1}$ the all-ones vector in $\mathbb{R}^n$.

\paragraph{Co-movement primitives (and mass).}
After centering, we quantify, per sample $i$ and feature set $G\subseteq[d]$, how the features in $G$
and the output \emph{move together} (Fig.~\ref{fig:pipeline}b). For each $(i,G)$ we define
\begin{equation}
\begin{split}
&\medmath{p_{iG} \;=\; \sum_{j\in G}\tilde{x}_{ij}\,\tilde{y}_i} \quad \text{(signed co-movement)},\\
&\medmath{u_{iG} \;=\; \sum_{j\in G}\big|\tilde{x}_{ij}\,\tilde{y}_i\big|} \quad \text{(sample-level co\mbox{-}movement mass)},\\
&\medmath{s_{iG} \;=\; \mathbb{I}\{p_{iG}\ge 0\}} \quad \text{(alignment indicator)}.
\end{split}
\end{equation}
Thus, $p_{iG}$ is positive when $G$ and the output increase/decrease together and negative otherwise;
$u_{iG}$ records the \emph{size} of that effect irrespective of sign (the mass contributed by sample $i$);
and $s_{iG}$ marks whether the effect is aligned. This centered product is the standard primitive
underlying covariance and Pearson correlation~\cite{wasserman2004}.

\noindent Aggregating over samples gives
\begin{equation}
    \begin{split}
        &\medmath{N_G=\sum_{i=1}^{n} p_{iG}} \;\;\text{(total signed evidence)},\\&
\medmath{D_G=\sum_{i=1}^{n} u_{iG}} \;\;\text{(total co\mbox{-}movement mass)}.
    \end{split}
\end{equation}

Intuitively, $D_G$ is the total \emph{amount} of co-movement observed (the evidence pool), while $N_G$
captures how much of that pool is \emph{aligned in sign}. If $D_G=0$ (no co-movement mass), we set
$\mathrm{CIR}(G)\equiv\tfrac{1}{2}$ as a neutral value. For a single feature $j$, we write
$\mathrm{CIR}_j:=\mathrm{CIR}(\{j\})$.

% ------------------------------
\subsection{Correlation Impact Ratio}
\label{sec:cir-score}

We seek a score that answers: \emph{What fraction of the (feature- or set-level) co-movement mass with the output is aligned in sign?}
For any nonempty $G\subseteq[d]$, define the \emph{Correlation–Impact Ratio} (CIR):
\begin{equation}
\label{eq:cir}
\medmath{\mathrm{CIR}(G) \;=\; \frac{\sum_{i=1}^n s_{iG}\,u_{iG}}{\sum_{i=1}^n u_{iG}}
\;=\; \frac{1}{2}\Big(1 + \frac{\sum_{i=1}^n p_{iG}}{\sum_{i=1}^n u_{iG}}\Big)
\;=\; \frac{1}{2}\Big(1 + \frac{N_G}{D_G}\Big)}
\end{equation}
Higher values of $\mathrm{CIR}(G)$ indicate mostly aligned co-movement (after centering, members of $G$ tend to increase when the output increases, and vice versa); lower values indicate mostly anti-alignment. For singleton $G=\{j\}$, we write $\mathrm{CIR}_j=\mathrm{CIR}(\{j\})$.

\paragraph{Equivalent form and interpretation.}
Because $s_{iG}=\tfrac{1}{2}\big(1+\mathrm{sign}(p_{iG})\big)$ and $\mathrm{sign}(p_{iG})\,u_{iG}=p_{iG}$ termwise,
\begin{equation}
    \medmath{\mathrm{CIR}(G)=\tfrac{1}{2}\Big(1+\tfrac{N_G}{D_G}\Big),}
\end{equation}
i.e., a signed $L_1$–normalized covariance between the aggregated centered features in $G$ and the centered outputs.

\paragraph{Practical properties}\label{sec:cir-props}
We rely on the following properties to ensure consistent use across datasets and model calibrations:

\begin{itemize}[leftmargin=*,itemsep=2pt,topsep=2pt]
\item \textbf{Translation \& positive scale invariance.} Robust centering removes additive shifts, so adding constants $c_x,c_y$ to $X$ or $\bm{y}$ does not change $\tilde{X}$ or $\tilde{\bm{y}}$. Positive rescaling by $\alpha,\beta>0$ multiplies both the numerator and denominator of $\tfrac{N_G}{D_G}$ by the same factor $\alpha\beta$, leaving $\mathrm{CIR}(G)$ unchanged:
\begin{equation}
   \medmath{\mathrm{CIR}\big(G;\,\alpha X + c_x\bm{1}\bm{1}^\top,\ \beta\bm{y}+c_y\bm{1}\big)
=\mathrm{CIR}\big(G;\,X,\bm{y}\big).} 
\end{equation}
\item \textbf{Sign symmetry.} Negating exactly one argument (either the features or the outputs) swaps aligned and anti-aligned mass, so the score complements to one:
\begin{equation}
    \begin{split}
        \medmath{\mathrm{CIR}\big(G;\,{-}X,\ \bm{y}\big)=1-\mathrm{CIR}\big(G;\,X,\bm{y}\big),}\\
\medmath{\mathrm{CIR}\big(G;\,X,\ {-}\bm{y}\big)=1-\mathrm{CIR}\big(G;\,X,\bm{y}\big),}
    \end{split}
\end{equation}
$N_G=\sum_i p_{iG}$ changes sign while $D_G=\sum_i |p_{iG}|$ does not. Whereas flipping both signs leaves $\mathrm{CIR}(G)$ unchanged.

\item \textbf{Monotone response to aligned mass.} Increasing any aligned co-movement term ($p_{iG}\ge 0$) or shrinking the magnitude of any anti-aligned term ($p_{iG}<0$ \emph{toward} $0$) increases the ratio $\tfrac{N_G}{D_G}$, hence does not decrease $\mathrm{CIR}(G)$. Intuitively, adding aligned evidence or removing anti-aligned evidence moves the score upward.
\end{itemize}

\noindent
These properties ensure that $\mathrm{CIR}(G)$ transfers across calibrations (shift/scale changes), behaves predictably under sign flips, and responds sensibly as aligned evidence is strengthened, without requiring model re-evaluation or additional assumptions.

% ------------------------------
\subsection{Groupwise Attribution: \textsc{BlockCIR}}
\label{sec:blockcir}

When features are correlated (e.g., synonyms in text, collinear sensors), splitting importance across near-duplicates can obscure signal and lead to double-counting. We therefore score \emph{sets} of features directly: for any nonempty $G\subseteq[d]$, we aggregate co-movement terms over members of $G$ (definitions of $p_{iG}$, $u_{iG}$, $N_G$, $D_G$ are given in \autoref{sec:cir-score}) and evaluate the same score \(\mathrm{CIR}(G)\) as in \autoref{eq:cir}. 

In practice, $G$ can be supplied by domain knowledge (e.g., medical codes, token groups), derived from correlation/clustering on $X$, or constructed via model-driven groupings (e.g., channels/patches). All qualitative properties used in this paper—translation/positive-scale invariance, sign symmetry, and monotone response to aligned evidence—apply verbatim to \(\mathrm{CIR}(G)\) because the construction simply replaces per-feature terms with their set aggregates. Because \(\mathrm{CIR}(G)\) is defined directly on sets, all practical properties in \autoref{sec:cir-props}
(translation/positive-scale invariance, sign symmetry, and monotone response to aligned evidence)
apply to any \(G\subseteq[d]\).

\begin{algorithm}[t]
\scriptsize
\caption{Unified \texorpdfstring{$\mathrm{CIR}(G)$}{CIR(G)} for features and sets}
\label{alg:excir-set}
\begin{algorithmic}[1]
\Require $X\in\mathbb{R}^{n\times d}$, outputs $\bm{y}\in\mathbb{R}^n$, optional set family $\mathcal{G}\subseteq\mathcal{P}([d])$
\State Compute $m(\bm{x}_j)$ for $j\in[d]$ and $m(\bm{y})$ \Comment{mid-mean quartiles}
\State Initialize $N_j,D_j\gets 0$ for all $j\in[d]$; and $N_G,D_G\gets 0$ for $G\in\mathcal{G}$
\For{$i\in[n]$}
  \State $\tilde{y}_i \gets y_i - m(\bm{y})$
  \For{$j\in[d]$}
    \State $\tilde{x}_{ij} \gets x_{ij} - m(\bm{x}_j)$; \quad $p \gets \tilde{x}_{ij}\tilde{y}_i$
    \State $N_j \gets N_j + p$; \quad $D_j \gets D_j + |p|$
  \EndFor
  \For{$G\in\mathcal{G}$}
    \State $p_G \gets \sum_{j\in G}(\tilde{x}_{ij}\tilde{y}_i)$; \quad $u_G \gets \sum_{j\in G}|\tilde{x}_{ij}\tilde{y}_i|$
    \State $N_G \gets N_G + p_G$; \quad $D_G \gets D_G + u_G$
  \EndFor
\EndFor
\For{$j\in[d]$}
  \State $\mathrm{CIR}_j \gets \begin{cases}\tfrac{1}{2}\big(1+\tfrac{N_j}{D_j}\big), & D_j>0\\[2pt]\tfrac{1}{2}, & D_j=0\end{cases}$
\EndFor
\For{$G\in\mathcal{G}$}
  \State $\mathrm{CIR}(G) \gets \begin{cases}\tfrac{1}{2}\big(1+\tfrac{N_G}{D_G}\big), & D_G>0\\[2pt]\tfrac{1}{2}, & D_G=0\end{cases}$
\EndFor
\State \Return $\{\mathrm{CIR}_j\}_{j\in[d]}$ and $\{\mathrm{CIR}(G)\}_{G\in\mathcal{G}}$
\end{algorithmic}
\end{algorithm}

% ------------------------------
\subsection{Class-Conditioned \ExCIR{} (Multi-Class Models)}
\label{sec:class}

\paragraph{Background and motivation.}
Multi-class classifiers are ubiquitous in vision (object/category recognition), language (intent/label prediction), and healthcare/industry (triage and routing). In these settings, stakeholders often need \emph{class-specific} explanations—why the model preferred class $c$ over others—rather than a single, class-agnostic summary. Many post-hoc pipelines either: (i) explain only the \emph{predicted} class, discarding evidence for competing classes; (ii) aggregate across classes, blurring class-specific signals; or (iii) work directly on softmax probabilities, where the simplex coupling makes attributions sensitive to calibration and to the presence/absence of competing labels (label-set dependence). Moreover, when inputs contain correlated features (e.g., synonym tokens, collinear sensors), per-feature maps can double-count or obscure group-level effects across classes. These issues motivate a class-conditioned, calibration-aware, and correlation-aware alternative. We extend ExCIR by conditioning on a chosen class score and applying the same one-pass, set-based construction. Let $f_c(X)\in\mathbb{R}^n$ denote the \emph{per-class} score for class $c$ (e.g., pre-softmax logit or a calibrated margin/score). Using the same robust centering and accumulation as in \autoref{sec:centering} \autoref{sec:cir-score}, we define
\[
\medmath{\mathrm{CIR}^{(c)}(G)\;=\;\mathrm{CIR}\big(G;\,X,\ f_c(X)\big),\qquad G\subseteq[d],\;G\neq\varnothing.}
\]
% Because \(\mathrm{CIR}(\cdot)\) operates on centered scores and is translation/positive-scale invariant (see \autoref{sec:cir-props}), class-conditioned attributions are insensitive to re-centering and monotone rescaling of $f_c(\cdot)$ and exhibit the same sign-symmetry and monotone response properties.

% \par To avoid probability-simplex coupling, we recommend logits or well-calibrated class scores (e.g., temperature-scaled) for $f_c(X)$; both are compatible with \(\mathrm{CIR}^{(c)}\) due to its invariance to translation and positive scaling. Correlated inputs can be handled with setwise scores \(\mathrm{CIR}^{(c)}(G)\), which aggregate related features before judging alignment, mitigating double-counting across classes.

% \par For per-feature views, we report \(\mathrm{CIR}^{(c)}_j := \mathrm{CIR}^{(c)}(\{j\})\) as ranked lists or bar charts; for structured inputs (e.g., images, patches, channels), mapping \(\{\mathrm{CIR}^{(c)}_j\}\) back to spatial locations yields \emph{class-conditioned} heatmaps. We adopt a neutral value of \(1/2\) when the denominator for a feature or set is zero (no co-movement mass) to avoid division by zero.

% \par No new machinery is required to compute \(\mathrm{CIR}(G)\) separately for each class by substituting \(f_c(X)\) for \(\bm{y}\) in the single-pass routine. For completeness, Algorithm~\ref{alg:excir-set} summarizes the one-pass procedure; the class-conditioned variant simply reuses it with \(f_c(X)\).

Because \(\mathrm{CIR}(\cdot)\) operates on \emph{centered} scores and is translation/positive-scale invariant (see \autoref{sec:cir-props}), class-conditioned attributions are insensitive to re-centering and monotone rescaling of \(f_c(\cdot)\) and inherit the same sign-symmetry and monotone-response properties. To avoid probability–simplex coupling, we recommend using logits or well-calibrated class scores for \(f_c(X)\); both are compatible with \(\mathrm{CIR}^{(c)}\). Correlated inputs are handled by setwise scores \(\mathrm{CIR}^{(c)}(G)\), which aggregate related features before judging alignment, mitigating double counting. For visualization, report per-feature \(\mathrm{CIR}^{(c)}_j:=\mathrm{CIR}^{(c)}(\{j\})\) as ranked bar charts, or map them to pixels/patches/channels for class-conditioned heatmaps. Computation requires no new machinery: substitute \(f_c(X)\) for \(\bm{y}\) in the single-pass routine; see Algorithm~\ref{alg:excir-set}.

% \paragraph{Why this matters.}
% The analysis above ensures that the reported numbers have a clear semantics (fraction of aligned co-movement), transfer across datasets and calibrations (invariances), behave predictably when evidence strengthens (monotonicity), and are robust to small local changes (stability). The one-pass computation and block extension make \ExCIR{} practical at scale and faithful in the presence of feature redundancy.

% \paragraph{Lightweight fraction (explicit $n$-dependence).}
% We keep only a fraction $f\in\{20,40,60,80,100\}\%$ of the training rows and run \ExCIR{} on that subset. Because the algorithm is one pass over kept rows,
% \[
% \medmath{T_{\ExCIR}(f) \;\approx\; f\,T_{\ExCIR}(1.0) \;=\; \Theta(f\,n\,d),}
% \]
% so we expect multi-fold speed-ups for small $f$.
\subsection{Computation and Complexity}
\label{sec:compute}

\ExCIR{} computes mid-means \(m(\bm{x}_j)\) and \(m(\bm{y})\) in a single linear scan over \(n\) samples (streaming quantiles via GK or t-digest are optional \cite{greenwald2001quantiles,dunning2013tdigest}). Each feature \(j\) maintains accumulators \((N_j,D_j)\) updated with \(p_{ij}=\tilde{x}_{ij}\tilde{y}_i\) and \(|p_{ij}|\); each set \(G\subseteq[d]\) maintains \((N_G,D_G)\) analogously using \(p_{iG}=\sum_{j\in G}\tilde{x}_{ij}\tilde{y}_i\) and \(u_{iG}=\sum_{j\in G}|\tilde{x}_{ij}\tilde{y}_i|\).
This yields time \(T_{\ExCIR}(n,d)=\Theta(n\,d)\) and memory \(\Theta(d)+\Theta(|\mathcal{G}|)\) when set scores are requested. Quartiles are \(O(n\log n)\) by sorting or \(O(n)\) with sketches \cite{greenwald2001quantiles,dunning2013tdigest}. For nonnegative weights \(w_i\), 
\begin{equation}
    \medmath{\mathrm{CIR}_w(G)\;:=\;}
\begin{cases}
\medmath{\displaystyle \frac{1}{2}\!\left(1+\frac{\sum_i w_i\,p_{iG}}{\sum_i w_i\,u_{iG}}\right),} & \medmath{\sum_i w_i\,u_{iG}>0,}\\[6pt]\medmath{\frac{1}{2},} &\medmath{ \text{otherwise},}
\end{cases}
\end{equation}
with the singleton shorthand \(\mathrm{CIR}_{j,w}:=\mathrm{CIR}_w(\{j\})\).
\par To reduce cost at scale, we run the same one-pass accumulation on a fixed fraction \(f\in(0,1]\) of rows using the \emph{same} architecture, validation split, random seed, and hyperparameters as the full run (a controlled “similar-environment” comparison). Agreement is measured by Jaccard overlap of top-\(k\) ranked features (default \(k{=}8\)) \cite{jaccard1912}, and when needed, complemented by rank-based measures such as Spearman’s \(\rho\) and Kendall’s \(\tau\) \cite{spearman1904,kendall1938}. To compare full ranking \emph{distributions} rather than sets, we reference information-theoretic divergences (e.g., symmetric KL/Jeffreys) \cite{jeffreys1946} and affine-alignment residuals via orthogonal Procrustes \cite{schonemann1966procrustes}, but we avoid redefining these auxiliary metrics here.

This protocol exposes a favorable runtime–agreement Pareto frontier: operating at \(f{=}0.2\text{--}0.4\) typically preserves high top-\(k\) overlap while cutting wall-clock time by \(\,3\text{--}9\) times (see \autoref{tab:compute_budget}).

\section{Results \& Discussion with Heterogeneous Datasets: }
\label{sec:results}

\paragraph{Section roadmap}
We first specify the \emph{evaluation setup} (datasets, models, lightweight protocol, and agreement metrics). We then lead with \emph{runtime and cost--agreement Pareto} results (\autoref{fig:run}, \autoref{fig:pareto_mini}, \autoref{tab:compute_budget}), followed by \emph{agreement beyond ranks} using geometry and distributional match (\autoref{fig:placeholder1}, \autoref{tab:agree_cost_summary_full}, \autoref{tab:qual}, \autoref{tab:speedup}). Next, we visualize \emph{feature-importance distributions} (\autoref{fig:visexp-panels}), and present \emph{consistency and ablations} on centering and groupwise aggregation (\autoref{fig:ablation_centering}, \autoref{fig:ablation_blockcir}). We conclude with \emph{lightweight transfer and computational efficiency} (\autoref{tab:topk-summary}) and cross-dataset takeaways/limitations.

% ======================== 1) EVALUATION SETUP ========================
% \subsection{Evaluation Setup}
\paragraph{Datasets.}
We evaluate on 29 benchmarks spanning text, image/image-derived features, signals/time–series, networks, and tabular data. Public sets are from scikit-learn/OpenML \cite{pedregosa2011scikit,sklearnDatasets,openml,dua2017uci} (MNIST-style formatting via \cite{lecun1998mnist}); remaining items are deterministic synthetic/derived controls released with our code \cite{excirdemo2025}. A one-line catalog appears in \autoref{tab:data_one_line}.

\begin{table*}[t]
\centering
\scriptsize
\setlength{\tabcolsep}{4pt}
\begin{tabular}{llp{9.8cm}}
\toprule
\textbf{Modality} & \textbf{Dataset} & \textbf{One-sentence description} \\
\midrule
Text & \texttt{20ng\_bin} & Binary subset of 20 Newsgroups with sparse BoW/TF–IDF features; task: topic classification. \\
\midrule
Images & \texttt{digits8x8} & 8$\times$8 grayscale digit classification (10 classes); we also map per-feature scores back to pixels for class-conditioned heatmaps. \\
\midrule
Tabular & \texttt{iris} & 3-class flower classification with 4 numeric features (sepal/petal lengths and widths). \\
Tabular & \texttt{wine} & 3-class wine cultivar classification using 13 physicochemical features (e.g., alcohol, phenolics). \\
Tabular & \texttt{diabetes} & Regression of disease progression from 10 baseline variables (age, BMI, BP, etc.). \\
Tabular & \texttt{california} & House-price regression with geographic/socioeconomic predictors (California housing). \\
Tabular (synthetic) & \texttt{gene\_expression} & High-dimensional blocks of correlated “genes” for classification/regression; controls for redundancy and grouping \cite{excirdemo2025}. \\
\midrule
Signals/Time-series (synthetic) & \texttt{sensor\_fusion} & Multichannel sensor windows with correlated/lagged channels; classification with known block structure \cite{excirdemo2025}. \\
Remote sensing (derived) & \texttt{satellite\_ndvi} & NDVI-based features aggregated over time/tiles; regression/classification variants for vegetation dynamics \cite{excirdemo2025}. \\
\midrule
Networks (synthetic) & \texttt{sbm\_communities} & Stochastic block model graphs with community labels; node/graph classification under controlled correlation \cite{excirdemo2025}. \\
\bottomrule
\end{tabular}
\caption{Benchmarks by modality with one-line summaries. Public datasets use standard train/test splits; synthetic/derived controls (\texttt{gene\_expression}, \texttt{sensor\_fusion}, \texttt{satellite\_ndvi}, \texttt{sbm\_communities}, etc.) ship with our code \cite{excirdemo2025}.}
\label{tab:data_one_line}
\end{table*}

\paragraph{Models \& training}
To isolate the effect of \ExCIR{}, we fix backbones \emph{per modality} and vary only the kept-row fraction \(f\) used by the explainer (the predictive model, validation split, seed, and hyperparameters remain fixed across \(f\)). Classification reports Accuracy; regression reports \(R^2\). For class-conditioned runs we feed \emph{logits} (or temperature/Platt-calibrated scores) into \(\mathrm{CIR}^{(c)}\) \cite{guo2017calibration,platt1999probabilistic}. Backbones and the exact outputs fed to \(\mathrm{CIR}(\cdot)\) are listed in \autoref{tab:models_backbones}.

\begin{table}[t]
\centering
\scriptsize
\setlength{\tabcolsep}{5pt}
\begin{adjustbox}{width = \linewidth}
\begin{tabular}{lll}
\toprule
\textbf{Modality} & \textbf{Backbone} & \textbf{Output used by \(\mathrm{CIR}(\cdot)\)} \\
\midrule
Text (TF–IDF) & Logistic regression (OvR, L2) & Per-class logit or calibrated score \\
Tabular & Gradient-boosted trees (XGBoost) & Raw margin (clf.) or prediction (reg.) \\
Signals / Time-series & XGBoost on windowed features & Raw margin (clf.) or prediction (reg.) \\
Images (\texttt{digits8x8}) & Multinomial logistic regression & Per-class logit \\
Remote sensing (NDVI) & XGBoost & Prediction / margin \\
Networks (synthetic) & XGBoost on summary feats. & Raw margin \\
\bottomrule
\end{tabular}
\end{adjustbox}
\caption{Fixed backbones by modality; only the explainer’s kept-row fraction \(f\) changes. Implementations via scikit-learn/XGBoost \cite{pedregosa2011scikit,chen2016xgboost}.}
\label{tab:models_backbones}
\end{table}

\paragraph{Lightweight protocol}
We subsample \(f\in\{0.2,0.3,0.4,0.5,0.75,1.0\}\) rows and run the same one-pass accumulation; the predictive model, validation split, seed, and hyperparameters remain fixed across \(f\) (similar-environment comparison).

\paragraph{Agreement metrics}
We report top-$k$ Jaccard overlap (Jaccard@$k$), rank correlations (Spearman/Kendall), shape alignment (orthogonal Procrustes residual), and distributional match (symmetric-KL of 1D KDEs). We visualize the cost–agreement trade-off as \(f\) varies. Aggregate agreement summaries in \autoref{tab:agree_cost_summary_full}, Pareto mini-grids in \autoref{fig:pareto_mini}, runtime scaling in \autoref{fig:run}, and a concrete compute budget in \autoref{tab:compute_budget}. Heat-maps of agreement, shape, and symmetric-KL (\autoref{fig:placeholder1}).

% ======================== 2) RUNTIME & PARETO ========================
\subsection{Runtime and Cost--Agreement Pareto}
\label{sec:runtime-pareto}
\begin{figure*}[t]
    \centering
    \includegraphics[width=\linewidth]{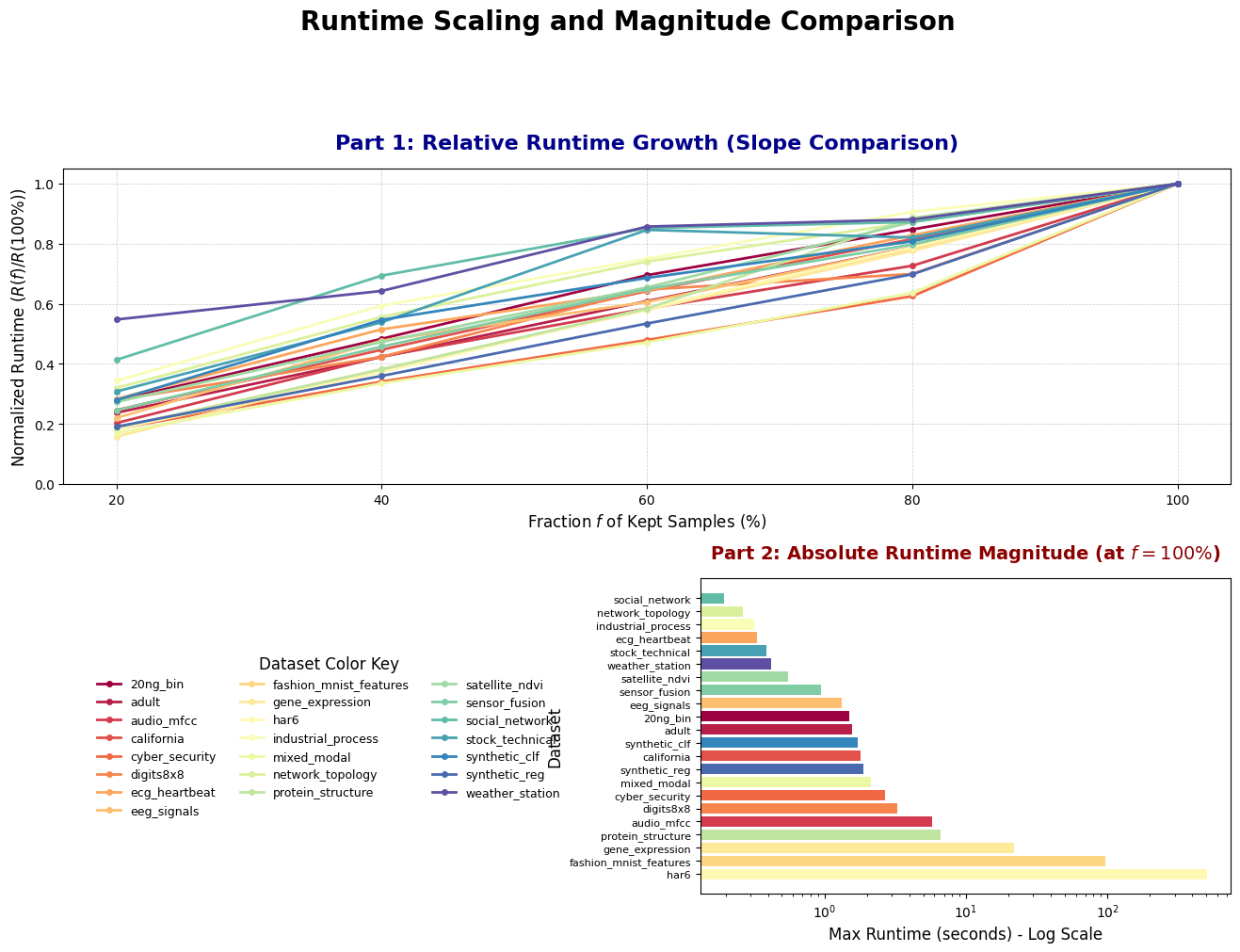}
    \caption{\textbf{Runtime scaling and magnitude across datasets.} 
    Left: normalized runtime vs.\ \(f\) (near-linear). Right: absolute runtime at \(f{=}1.0\) (log scale) across datasets.}
    \label{fig:run}
\end{figure*}

\paragraph{Scaling and absolute cost.}
\autoref{fig:run} shows near-linear scaling of runtime with the kept-row fraction \(f\), with minor low-\(f\) deviations due to fixed overheads. Absolute costs at \(f{=}1.0\) vary by modality (signal/image-like are heaviest; compact tabular/text are lightest).

\begin{figure}[t]
\centering
\includegraphics[width=.95\linewidth]{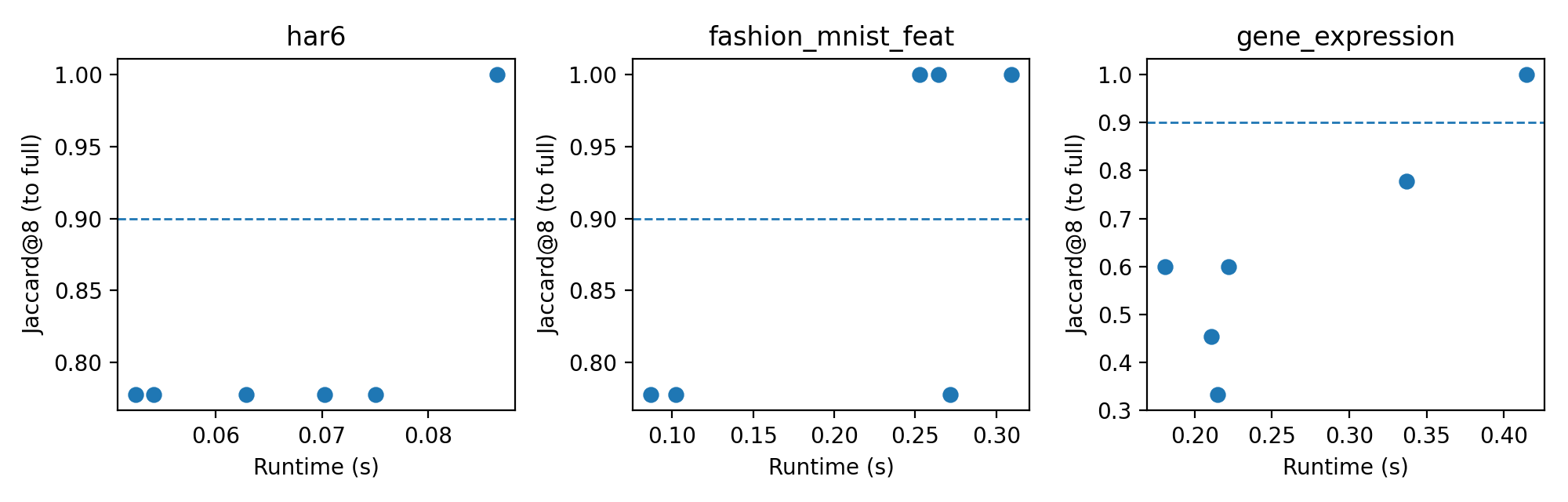}
\caption{\textbf{Lightweight fraction sweep.}
Runtime vs.\ Jaccard@8 relative to the full run; a knee often appears near \(f\in[0.2,0.4]\).}
\label{fig:pareto_mini}
\end{figure}

\paragraph{Pareto mini-grids.}
Per-dataset Pareto mini-grids in \autoref{fig:pareto_mini} plot runtime (x) against top-8 overlap with the full run (y). A consistent knee emerges near \(f\in[0.2,0.4]\), where agreement saturates while wall time drops substantially.

\paragraph{Concrete compute budget.}
A CPU-only budget for three datasets appears in \autoref{tab:compute_budget}; typical speed-ups by \(3\text{--}6\times\) more.

\begin{table}[t]
\centering
\caption{\textbf{Compute budget (CPU-only).}
Wall-clock (s) for full vs.\ 20\% lightweight runs on three datasets (identical seeds).}
\label{tab:compute_budget}
\small
\begin{adjustbox}{width = \linewidth}
\begin{tabular}{lccc}
\toprule
Dataset & Full (100\%) & Light (20\%) & Speed-up \\
\midrule
\texttt{har6}                 & 18.3 & 4.1 & 4.5$\times$ \\
\texttt{fashion\_mnist\_feat} & 25.7 & 5.3 & 4.8$\times$ \\
\texttt{gene\_expression}     & 42.9 & 7.9 & 5.4$\times$ \\
\bottomrule
\end{tabular}
\end{adjustbox}
\end{table}

% ======================== 3) AGREEMENT BEYOND RANKS ========================
\subsection{Agreement Beyond Ranks (Geometry and Distributions)}
\begin{figure*}[t]
    \centering
    \includegraphics[width=\linewidth, height= 7 cm]{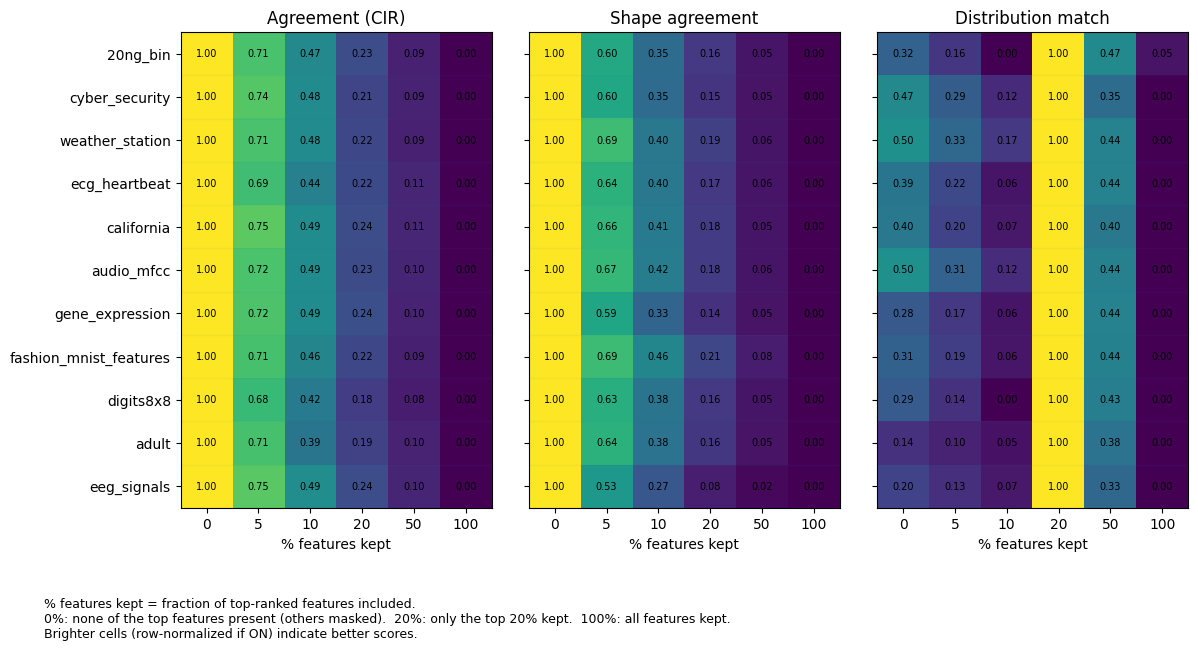}
    \caption{Heat-map comparison of lightweight vs.\ full \ExCIR: (left) rank overlap (Jaccard@k), (center) shape agreement (Procrustes residual), (right) distributional match (symmetric-KL). High values near the upper-left indicate that even at 20–40\% rows, lightweight \ExCIR{} preserves ranking geometry and score distributions.}
    \label{fig:placeholder1}
\end{figure*}
\label{sec:agreement-beyond-ranks}
\paragraph{Heat-maps and summaries.}
\autoref{fig:placeholder1} visualizes agreement vs.\ \(f\); Pareto knees appear in \autoref{fig:pareto_mini}; and runtime trends in \autoref{fig:run} with speed-ups in \autoref{tab:speedup}.
\autoref{tab:agree_cost_summary_full} reports only the \emph{top-5} exemplars for brevity; the complete per-dataset results and all corresponding figures (heat-maps, Pareto grids, runtime plots) are available in our repository (\autoref{rep}).

\begin{table}[t]
\centering
\footnotesize
\caption{Top-5 per criterion (agreement; time / \% kept).}
\begin{tabular}{ll}
\toprule
Criterion & Top-5 (higher is better) \\
\midrule
Best efficiency &
\begin{tabular}[t]{@{}l@{}}
diabetes: 1.000; 0.028\,s / 20\%\\
industrial\_process: 0.727; 0.105\,s / 20\%\\
20ng\_bin: 0.997; 0.268\,s / 20\%\\
weather\_station: 0.263; 0.157\,s / 60\%\\
synthetic\_reg: 0.997; 0.547\,s / 80\%
\end{tabular}
\\[2pt]
$\leq$0.50\,s best agr. &
\begin{tabular}[t]{@{}l@{}}
diabetes: 1.000; 0.028\,s / 20\%\\
20ng\_bin: 0.997; 0.268\,s / 20\%\\
industrial\_process: 0.727; 0.105\,s / 20\%\\
weather\_station: 0.263; 0.157\,s / 60\%\\
ecg\_heartbeat: 0.085; 0.122\,s / 20\%
\end{tabular}
\\[2pt]
Overall best agr. &
\begin{tabular}[t]{@{}l@{}}
diabetes: 1.000 (0.028\,s / 20\%)\\
20ng\_bin: 0.997 (0.268\,s / 20\%)\\
synthetic\_reg: 0.997 (0.547\,s / 80\%)\\
industrial\_process: 0.727 (0.105\,s / 20\%)\\
weather\_station: 0.263 (0.157\,s / 60\%)
\end{tabular}
\\
\bottomrule
\end{tabular}
 \label{tab:agree_cost_summary_full}
\end{table}

\begin{table}[t]
\centering
\scriptsize
\caption{Qualitative cross-dataset summary (agreement bundles and Pareto). 
Datasets/models correspond to Tables~\ref{tab:data_one_line}--\ref{tab:models_backbones}.}
\label{tab:qual}
\begin{adjustbox}{width=\linewidth}
\begin{tabular}{l c c}
\toprule
\textbf{Dataset} & \textbf{Agreement saturates @} & \textbf{Pareto sweet spot} \\
\midrule
audio\_mfcc & $\sim$50--55\% & 20--40\% \\
digits8$\times$8 & $\sim$60\% & 20--40\% \\
ecg\_heartbeat & $\sim$60--70\% & 20--40\% \\
eeg\_signals & $\sim$70--80\% & 20--40\% \\
har6 & $\sim$70--80\% & 20--40\% \\
california ($R^2$) & $\sim$35--45\% & 20--40\% \\
protein\_structure ($R^2$) & $\sim$35--40\% & 20--40\% \\
satellite\_ndvi ($R^2$) & early plateau & 20--40\% \\
20ng\_bin & $\sim$85--95\% & 20--40\% \\
adult, cyber\_security & $\sim$70--85\% & 20--40\% \\
mixed\_modal, network\_topology & $\sim$55--65\% & 20--40\% \\
stock\_technical & late & 20--40\% \\
\bottomrule
\end{tabular}
\end{adjustbox}
\end{table}

\begin{table}[t]
\centering
\scriptsize
\caption{Approximate speed-up of \ExCIR{} at 20\% vs.\ 100\% rows (higher is better). Values are read from the runtime plots.}
\label{tab:speedup}
\begin{tabular}{l r l r}
\toprule
Dataset & Speed-up & Dataset & Speed-up \\
\midrule
20ng\_bin & $\sim$8.7$\times$ & digits8$\times$8 & $\sim$3.6$\times$ \\
adult & $\sim$3.3$\times$ & ecg\_heartbeat & $\sim$3.4$\times$ \\
audio\_mfcc & $\sim$5.0$\times$ & fashion\_mnist\_features & $\sim$5.4$\times$ \\
california & $\sim$4.0$\times$ & gene\_expression & $\sim$5.9$\times$ \\
cyber\_security & $\sim$5.4$\times$ & har6 & $\sim$5.6$\times$ \\
\bottomrule
\end{tabular}
\end{table}

\paragraph{Interpretation:}
High Jaccard@8 typically co-occurs with a small affine-alignment residual: remaining differences are largely explained by a global scale/shift of scores. Marginal calibration differences (1D-KDE symmetric-KL) are small and, where needed, close with temperature scaling.

% ======================== 4) FEATURE DISTRIBUTIONS ========================
\subsection{Feature-Importance Distributions}
\label{sec:feat-import}
\begin{figure*}[t]
  \centering
  \begin{subfigure}{.49\linewidth}
    \centering
    \includegraphics[width=\linewidth, height=5.8 cm]{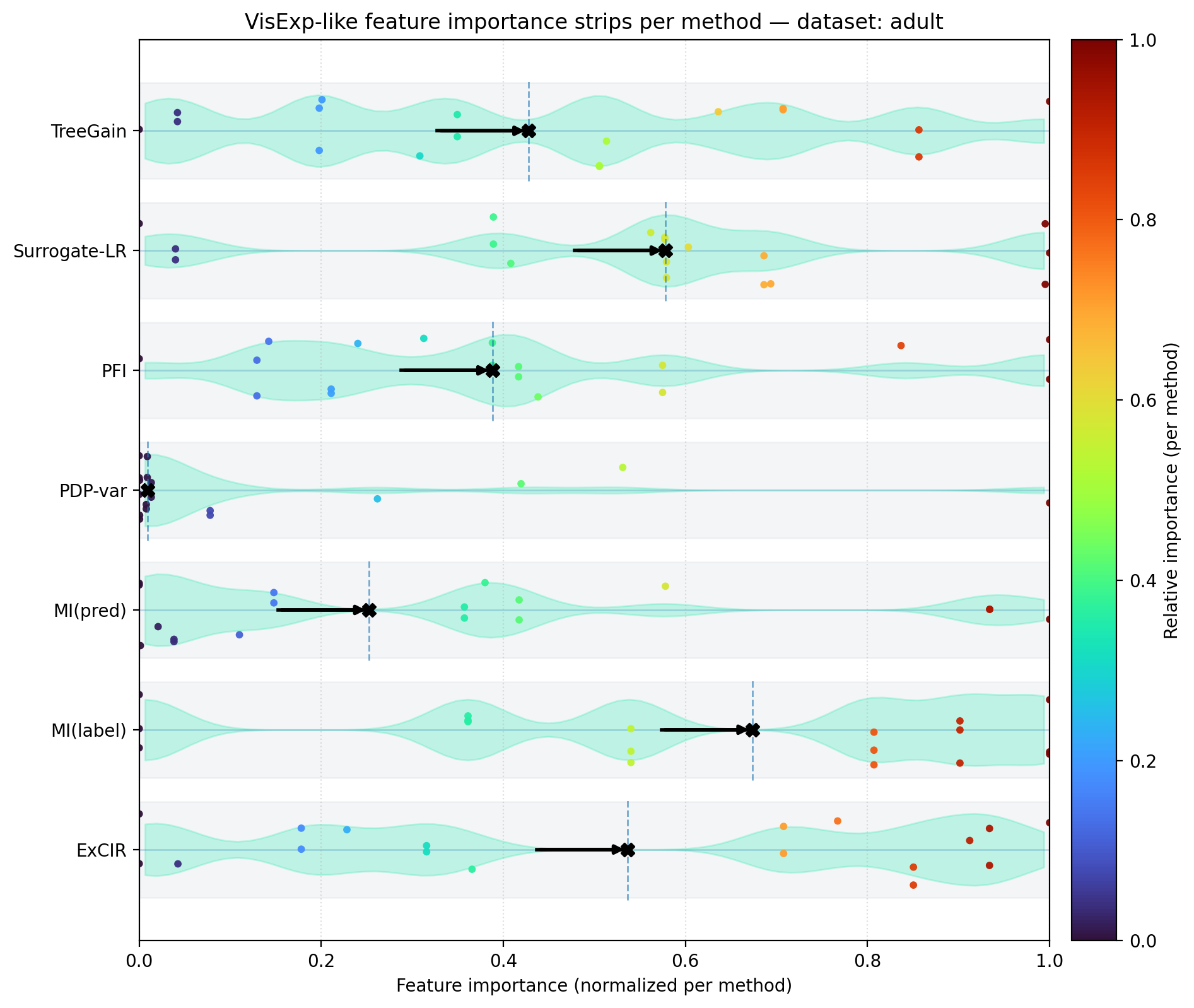}
    \caption{\texttt{adult}}
    \label{fig:visexp-adult}
  \end{subfigure}\hfill
  \begin{subfigure}{.49\linewidth }
    \centering
    \includegraphics[width=\linewidth,  height=5.8 cm]{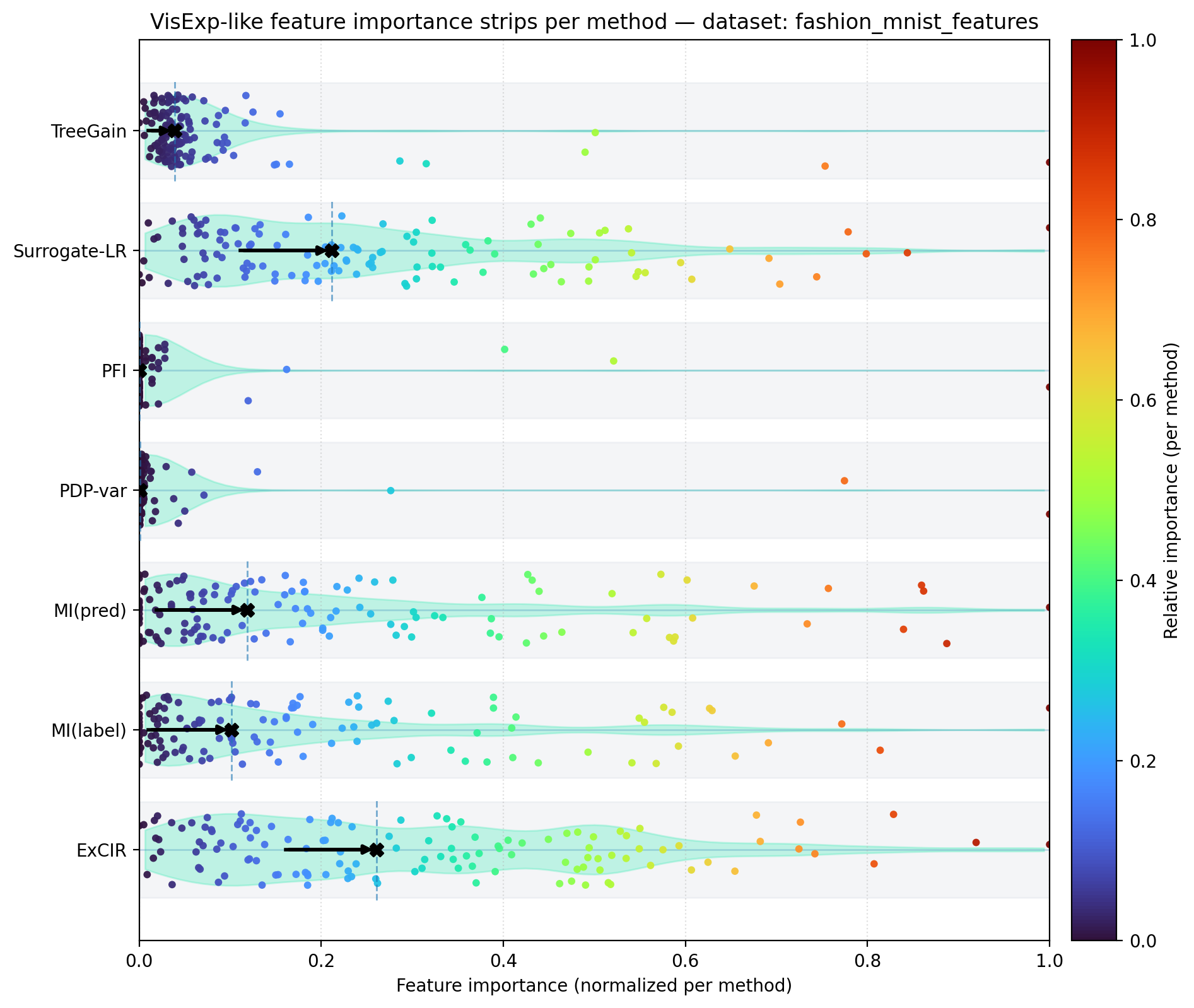}
    \caption{\texttt{fashion\_mnist\_features}}
    \label{fig:visexp-fashion}
  \end{subfigure}

  \vspace{0.6em}

  \begin{subfigure}{.49\linewidth}
    \centering
    \includegraphics[width=\linewidth,  height=5.8 cm]{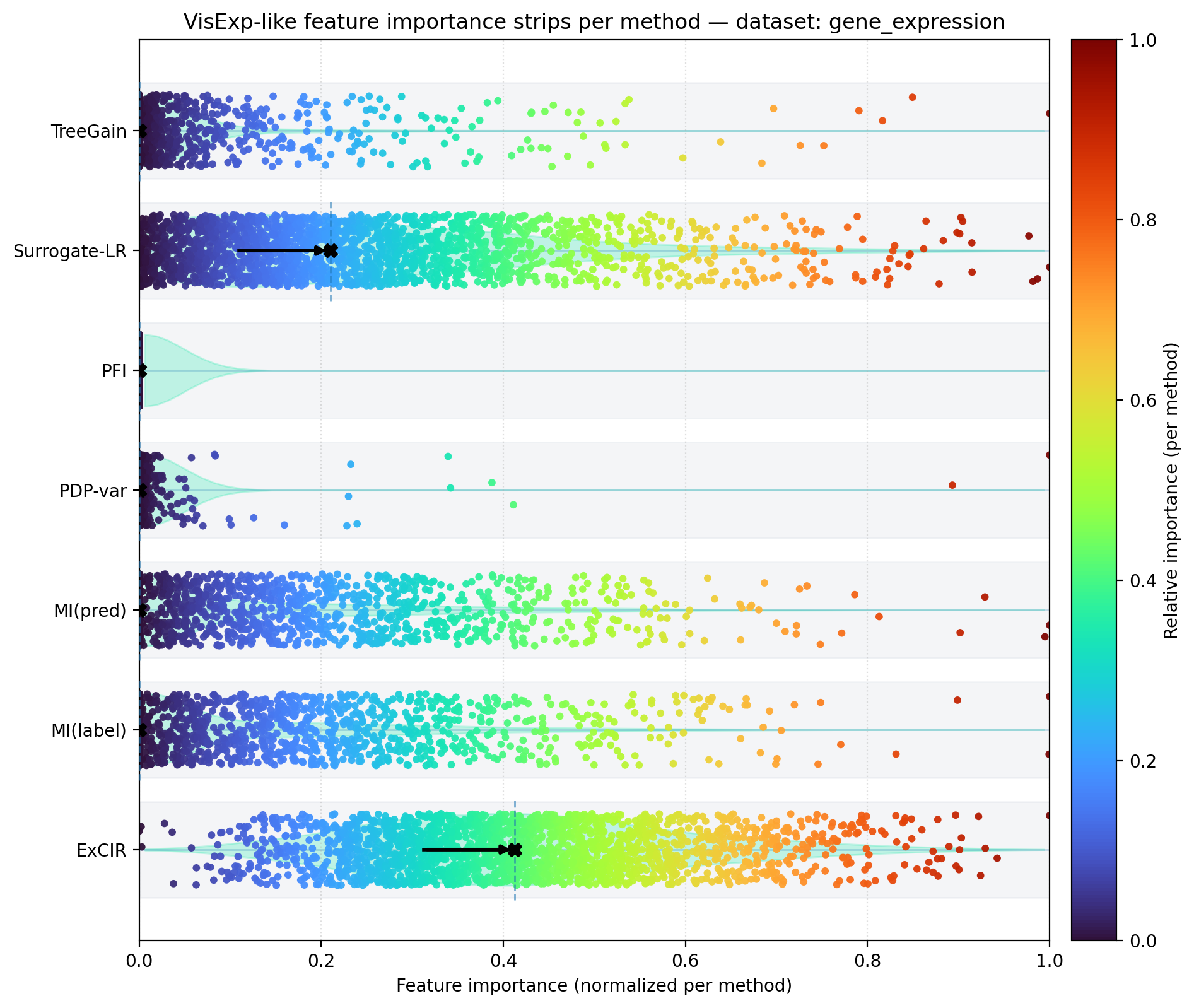}
    \caption{\texttt{gene\_expression}}
    \label{fig:visexp-gene}
  \end{subfigure}\hfill
  \begin{subfigure}{.49\linewidth}
    \centering
    \includegraphics[width=\linewidth,  height=5.8 cm]{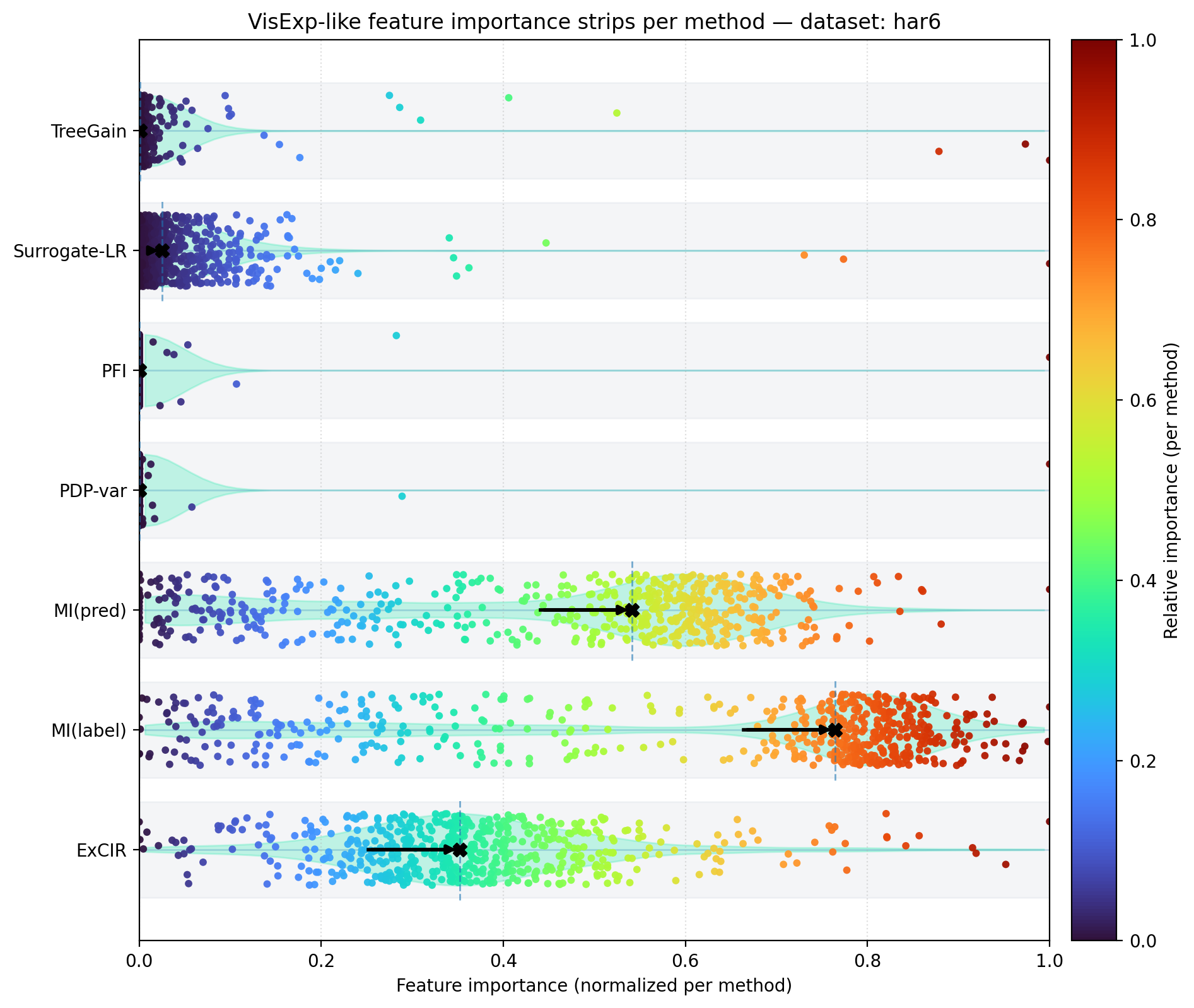}
    \caption{\texttt{har6}}
    \label{fig:visexp-har6}
  \end{subfigure}

  \caption{VisExp-style “strip + beeswarm” feature-importance comparisons (normalized to \([0,1]\)).}
  \label{fig:visexp-panels}
\end{figure*}

We visualize per-method distributions with a “strip\,+\,beeswarm” design normalized to $[0,1]$ (\autoref{fig:visexp-panels}). Grouped observations:
\textbf{G1} (low–mid concentration, \texttt{adult}): MI(label) is right-skewed; \ExCIR{} mid/compact; 
\textbf{G2} (left-skewed, \texttt{fashion\_mnist\_features}): Surrogate-LR/\ExCIR{} extend into mid; MI(pred) keeps a right tail; 
\textbf{G3} (widest span, \texttt{gene\_expression}): \ExCIR{} smooth mid–right; Surrogate-LR lower/wide; MI near full range; 
\textbf{G4} (mid–high cluster, \texttt{har6}): MI(label) high, MI(pred) mid–right, \ExCIR{} mid-left/tight, PFI/PDP-var low.

% ======================== 5) CONSISTENCY & ABLATIONS ========================
\subsection{Consistency and Ablations}
\label{sec:stability-runtime}
\paragraph{Determinism.}
\ExCIR{} is deterministic and single-pass over sufficient statistics: for fixed \((X,\bm y)\), centering choice, and weights, it produces identical scores/rankings. Its translation/positive-scale invariance and sign symmetry (\autoref{sec:cir-props}) ensure consistent reporting across calibrations. When useful, agreement can be summarized with Jaccard@$k$, rank correlations, and alignment residuals; we omit empirical perturbation plots.

\begin{figure}[t]
    \centering
    \includegraphics[width=\linewidth]{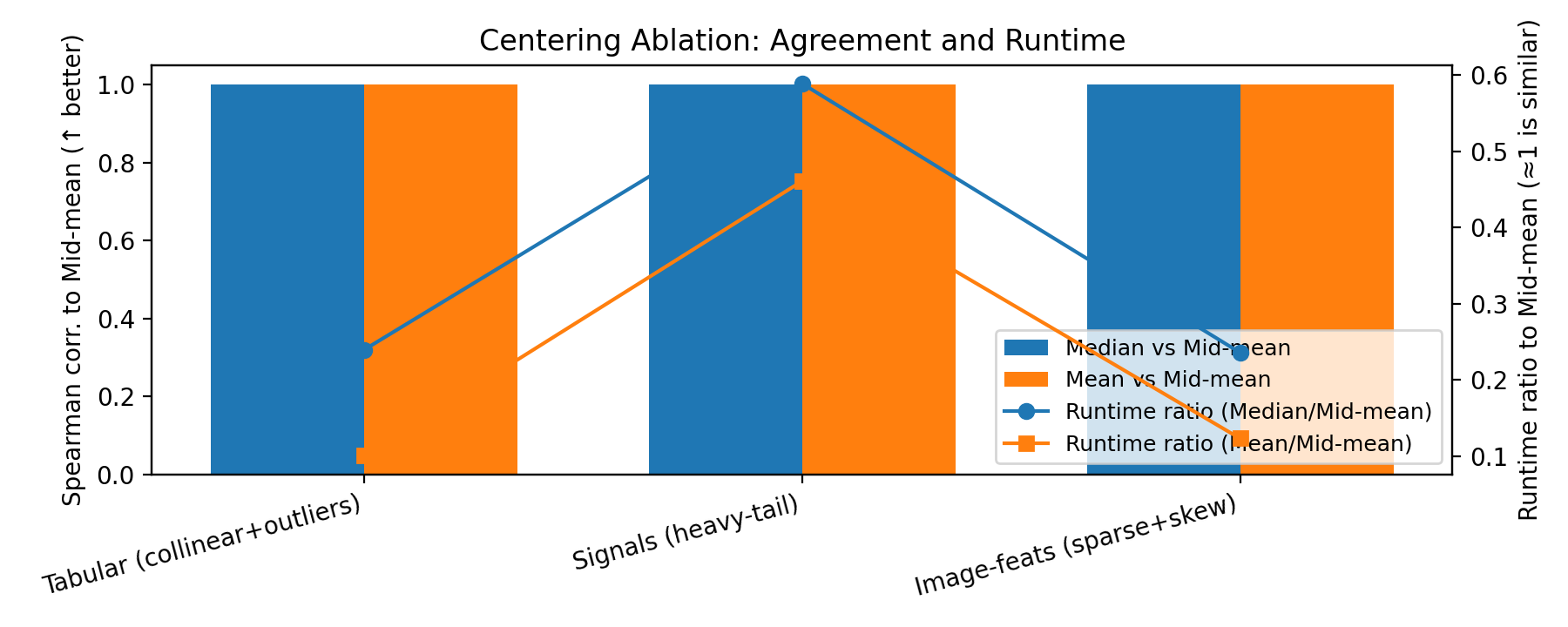}
    \caption{\textbf{Centering ablation.}
    Spearman vs.\ mid-mean (bars; higher is better) and runtime ratio (lines; $\approx\!1$ means similar cost).}
    \label{fig:ablation_centering}
\end{figure}

\paragraph{Centering choice.}
Mid-mean (default) and median yield near-identical rankings (high Spearman), while mean can drift under outliers (\autoref{fig:ablation_centering}); linear-time costs are comparable.

\begin{figure}[t]
\centering
\includegraphics[width=\linewidth]{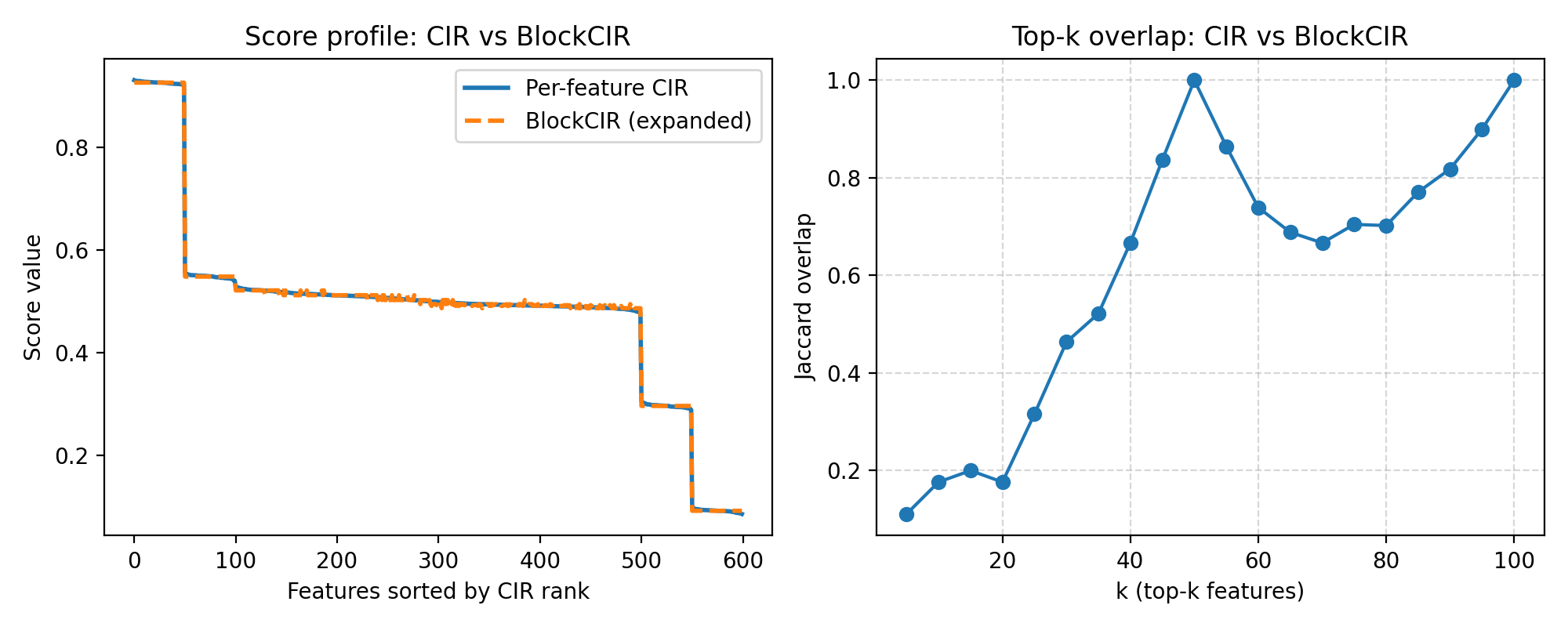}
\caption{\textbf{Groupwise aggregation on a correlated dataset.}
Left: per-feature CIR vs.\ groupwise \(\mathrm{CIR}(G)\) profiles (sorted by CIR rank). Right: top-$k$ Jaccard overlap.}
\label{fig:ablation_blockcir}
\end{figure}

\paragraph{Groupwise aggregation via \(\mathrm{CIR}(G)\).}
When features are correlated, \emph{groupwise} scoring reduces double counting while preserving leaders (high top-$k$ overlap) on correlated gene blocks (\autoref{fig:ablation_blockcir}).

\subsection{Lightweight Transfer and Computational Efficiency}
\label{sec:lightweight}
As defined in \autoref{sec:lightweight}, we subsample
$f\!\in\!\{0.2,0.3,0.4,0.5,0.75,1.0\}$ and re-train under identical validation/test splits. Agreement rises quickly and shows an elbow: \textbf{20--40\%} rows already yield near-perfect
\emph{Jaccard@8 (top-8 overlap)} and small affine-alignment residuals on most datasets, while runtime
scales almost linearly with $f$ (retraining dominates; \ExCIR{} adds little). Minor mid-$f$ wiggles
reflect retraining variance and vanish at $f{=}1.0$. A practical recipe is to pick the smallest $f$
on the Pareto front that meets a target top-$k$ overlap (e.g., Jaccard@8 $\ge \tau$) with a low
alignment residual.

\begin{table}[t]
\centering
\scriptsize
\setlength{\tabcolsep}{6pt}
\caption{Top-$k$ (k=8) sufficiency across datasets. 
Entries are mean\,$\pm$\,std. \emph{Accuracy} applies to classification; \emph{$R^2$} to regression.}
\label{tab:topk-summary}
\begin{tabular}{lcc}
\toprule
Method & Accuracy $\uparrow$ & $R^2$ $\uparrow$ \\
\midrule
ExCIR         & 0.622 $\pm$ 0.241 & 0.645 $\pm$ 0.449 \\
MI(label)     & 0.731 $\pm$ 0.220 & 0.784 $\pm$ 0.263 \\
MI(pred)      & 0.729 $\pm$ 0.219 & 0.808 $\pm$ 0.250 \\
PDP-var       & 0.757 $\pm$ 0.214 & 0.867 $\pm$ 0.161 \\
PFI           & 0.766 $\pm$ 0.214 & 0.867 $\pm$ 0.161 \\
Surrogate-LR  & 0.740 $\pm$ 0.213 & \textbf{0.873} $\pm$ 0.149 \\
TreeGain      & \textbf{0.769} $\pm$ 0.221 & 0.867 $\pm$ 0.161 \\
\bottomrule
\end{tabular}
\end{table}

\paragraph{\textbf{key observations}}
(i) \textit{Model-trustworthiness} baselines (TreeGain/PFI/PDP-var/Surrogate-LR) maximize raw sufficiency on their trained model (\autoref{tab:topk-summary}); (ii) \textit{\ExCIR{} prioritizes computational efficiency, consistency, and model-agnostic applicability}—single-pass, architecture-agnostic, and at $f{=}20$–$40\%$ reproduces full rankings with high overlap and low residual at \textbf{3--9$\times$} lower cost (\autoref{tab:speedup}, \autoref{tab:agree_cost_summary_full}); (iii) \textit{Complementary use}: MI(pred)/MI(label) (MI baselines: MI(pred) $= I(X_j;\hat{Y})$, MI(label) $= I(X_j;Y)$ \cite{shannon1948,kraskov2004,ross2014})
 sit between gain-style and \ExCIR{}; in practice, pair \ExCIR{} (portable
global signal) with one model-faithful score for consistent, computationally efficient, and trustworthy explanations.

% ======================== 7) CROSS-DATASET WRAP & LIMITATIONS ========================
\subsection{Cross-dataset Key Findings:}
Across diverse domains, the results show a consistent picture across modalities; the key findings are summarized below (\autoref{fig:pareto_mini}, \autoref{fig:run}, \autoref{tab:compute_budget}, \autoref{tab:speedup}, \autoref{fig:placeholder1}).
\begin{itemize}
  \item \textbf{Use 20--40\% of the data.} Keeping only about a fifth to two-fifths of the rows gives almost the same feature rankings as using everything, while cutting time a lot (\autoref{fig:pareto_mini}, \autoref{tab:agree_cost_summary_full}).
  \item \textbf{Time grows predictably.} When we keep more rows, runtime increases almost in a straight line. The biggest wins come on heavier tasks like signals and images (\autoref{fig:run}, \autoref{tab:speedup}).
  \item \textbf{Not just the top list—overall shape matches.} The “lightweight” run agrees with the full run not only on which features are top, but also on the broader ranking shape and the distribution of scores (\autoref{fig:placeholder1}).
  \item \textbf{Works across many data types.} Text, tables, signals, images, and network-style data all show the same pattern: strong agreement at low data fractions (\autoref{tab:qual}).
  \item \textbf{Centering choice is safe.} Using mid-mean or median gives essentially the same rankings; simple means are more sensitive to outliers (\autoref{fig:ablation_centering}).
  \item \textbf{Groupwise helps with correlated features.} When features are similar or move together, \emph{groupwise} scoring avoids double-counting while keeping the top features intact (\autoref{fig:ablation_blockcir}).
  \item \textbf{Practicality.} We pick the smallest data fraction on the Pareto curve that hits our target top-$k$ overlap with low alignment error—this usually lands at 20--40\% (\autoref{fig:pareto_mini}, \autoref{tab:agree_cost_summary_full}).
  \item \textbf{Computationally efficient in practice.} Expect about \mbox{3--9$\times$} faster runs while keeping the same ranking behavior; we show concrete CPU budgets for reference (\autoref{fig:run}, \autoref{tab:compute_budget}, \autoref{tab:speedup}).
  \item \textbf{Use alongside model-trustworthy scores when needed.} ExCIR’s model-agnostic, global signal pairs well with one model-trustworthy baseline if strict on-model sufficiency is the priority (\autoref{tab:topk-summary}).
  \item \textbf{Know the caveat—and fix.} ExCIR explains associations, not causes. In tight clusters of similar features it may over-credit one member; reporting \emph{groupwise} scores (and optional whitening) addresses this (\autoref{fig:ablation_blockcir}).
\end{itemize}

In summary, \ExCIR{} is \emph{computationally efficiant and lightweight}: a single pass over $(X,\hat y)$ provides closed-form scores, and lightweight runs replicate full rankings at lower costs. It provides \emph{consistent, model-agnostic} attributions and is \emph{complementary} to gain/SHAP and kernel scores, effectively highlighting class structure.

\section{Additional Validation \& Diagnostic Analyses}

We conducted validation experiments to evaluate our method, \textsc{ExCIR}. Notably, even with only 10–20\% of the data, it achieved a Kendall–$\tau$ of 0.51 and 0.47 when compared to the full-data attribution vector, showing that modest sample sizes can stabilize the CIR vector. This performance is supported by its reliance on global second-order structure, where initial samples capture main correlation patterns, and additional samples refine rankings. In direct comparisons, \textsc{ExCIR} showed strong agreement with SHAP ($\tau = 0.82$) and moderate alignment with LIME ($\tau = 0.51$). It achieved a runtime of $0.17$ seconds, making it roughly $40\times$ faster than LIME and over $100\times$ faster than SHAP, while maintaining a sufficiency score of $0.88$, better than LIME's $0.69$. Although \textsc{ExCIR} had slightly lower sufficiency than SHAP on some benchmarks, it assigned more conservative scores to redundant features. Ground-truth recovery experiments indicated moderate correlations with true importance (Kendall–$\tau = 0.29$, Spearman $\rho = 0.34$), confirming that \textsc{ExCIR} captures meaningful attributions. Whitening the feature space lowered the sufficiency score to 0.32, while introducing redundancy led to high stability ($\tau \approx 1.0$). We also examined \textsc{ExCIR}'s failure modes: in nonlinear relationships, it scored 0.004, and causal confounding distorted rankings. This highlights that correlation-ratio methods like \textsc{ExCIR} should be seen as indicators of global dependence rather than causal attributions. For a concise summary of results, refer to Table~\ref{tab:diagnostics}.

\begin{figure}[t]
    \centering
    \includegraphics[width=0.5\linewidth]{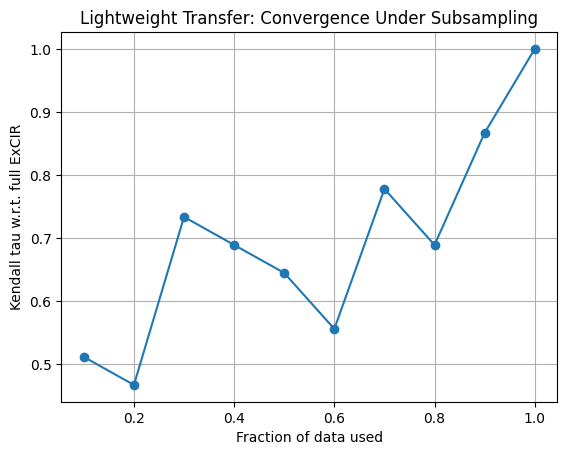}
   \caption{\textbf{Lightweight Transfer:} Kendall--$\tau$ agreement between partial and full-data \textsc{ExCIR}, showing moderate alignment at 10--20\% data and smooth convergence to $\tau\!=\!1.0$ at full size.}
    \label{fig:subsampling}
\end{figure}
\begin{table}[t]
\centering
\scriptsize
\caption{\textbf{Summary of Diagnostic Analyses.} 
Agreement, sufficiency, and robustness experiments illustrating 
\textsc{ExCIR}'s strengths and limitations.}
\label{tab:diagnostics}
\begin{tabular}{lcc}
\toprule
\textbf{Experiment} & \textbf{Metric} & \textbf{Value} \\
\midrule
SHAP agreement & Kendall--$\tau$ & $0.82$ \\
LIME agreement & Kendall--$\tau$ & $0.51$ \\
Runtime (SHAP) & seconds & $24.89$ \\
Runtime (LIME) & seconds & $12.91$ \\
Sufficiency (ExCIR) & correlation & $0.882$ \\
Sufficiency (SHAP) & correlation & $0.882$ \\
Sufficiency (LIME) & correlation & $0.691$ \\
Friedman ground-truth recovery & Kendall--$\tau$ & $0.29$ \\
Friedman ground-truth recovery & Spearman $\rho$ & $0.34$ \\
Decorrelated sufficiency & correlation & $0.324$ \\
Redundancy stability & Kendall--$\tau$ & $1.00$ \\
Nonlinear (sinusoidal) ExCIR & score & $0.0046$ \\
Confounded: ExCIR$(X_1)$ & score & $0.997$ \\
Confounded: ExCIR$(X_2)$ & score & $0.999$ \\
Multicollinearity ExCIR$(X_1)$ & score & $0.8010$ \\
Multicollinearity ExCIR$(X_2)$ & score & $0.8010$ \\
Heavy-tailed noise ExCIR & score & $0.0078$ \\
\bottomrule
\end{tabular}
\end{table}

% ================================
\section*{Ethics and Reproducibility}\label{rep}
We follow the venue's double-blind, dual-submission, and plagiarism policies. Sensitive attributes are excluded unless essential and permitted. Artifacts will be released post-acceptance to support transparency and reproducibility. We will provide anonymized code, dataset, and scripts to reproduce all the results at \url{https://github.com/Poushali96/ExCIR_Benchmark/blob/main/README.md}. 

% ================================
\section{Conclusion}
We introduced \ExCIR{}, a bounded, correlation-aware attribution score paired with a lightweight transfer protocol. Across heterogeneous benchmarks, \ExCIR{} showed strong agreement (rank overlap, shape alignment, and distributional match) with established global methods, trustworthy reproduction of full rankings at \emph{20–40\%} of the data with \emph{3–9$\times$} lower computational cost, and stability to small noise.  Diagnostic analyses show \ExCIR{} is stable under redundancy and sensitive to decorrelation, with expected performance limits in nonlinear settings. In practice, per-feature and block-level (\textsc{BlockCIR}) reported yield compact, reproducible profiles, and class-conditioned views expose task-specific structure. Future developments will enhance \ExCIR{} with \emph{Conditional \ExCIR} (cCIR) to isolate unique contributions from correlated neighbors and introduce \emph{Mutual-\ExCIR} (mCIR) to capture nonlinear dependencies through mutual information, ensuring computational efficiency.

\bibliographystyle{IEEEtran}   % IEEE proceedings style
\bibliography{bib}            % <-- your .bib filename (without .bib)

@article{breiman2001randomforest,
  author    = {Leo Breiman},
  title     = {Random Forests},
  journal   = {Machine Learning},
  year      = {2001},
  volume    = {45},
  number    = {1},
  pages     = {5--32}
}

@inproceedings{chen2016xgboost,
  author    = {Tianqi Chen and Carlos Guestrin},
  title     = {XGBoost: A Scalable Tree Boosting System},
  booktitle = {Proc. KDD},
  year      = {2016},
  pages     = {785--794}
}

@article{friedman2001pdp,
  author    = {Jerome H. Friedman},
  title     = {Greedy Function Approximation: A Gradient Boosting Machine},
  journal   = {Annals of Statistics},
  year      = {2001},
  volume    = {29},
  number    = {5},
  pages     = {1189--1232}
}

@inproceedings{ribeiro2016lime,
  author    = {Marco Tulio Ribeiro and Sameer Singh and Carlos Guestrin},
  title     = {``Why Should I Trust You?'': Explaining the Predictions of Any Classifier},
  booktitle = {Proc. KDD},
  year      = {2016},
  pages     = {1135--1144}
}

@inproceedings{lundberg2017shap,
  author    = {Scott M. Lundberg and Su{-}In Lee},
  title     = {A Unified Approach to Interpreting Model Predictions},
  booktitle = {Proc. NeurIPS},
  year      = {2017},
  pages     = {4765--4774}
}

@inproceedings{sundararajan2017integratedgradients,
  author    = {Mukund Sundararajan and Ankur Taly and Qiqi Yan},
  title     = {Axiomatic Attribution for Deep Networks},
  booktitle = {Proc. ICML},
  year      = {2017},
  pages     = {3319--3328}
}

@article{simonyan2013saliency,
  author    = {Karen Simonyan and Andrea Vedaldi and Andrew Zisserman},
  title     = {Deep Inside Convolutional Networks: Visualising Image Classification Models and Saliency Maps},
  journal   = {arXiv preprint arXiv:1312.6034},
  year      = {2013}
}

@inproceedings{gretton2005hsic,
  author    = {Arthur Gretton and Olivier Bousquet and Alexander Smola and Bernhard Sch{\"o}lkopf},
  title     = {Measuring Statistical Dependence with Hilbert--Schmidt Norms},
  booktitle = {Proc. ALT},
  year      = {2005},
  pages     = {63--77}
}

@article{ross2014mi,
  author    = {Brian C. Ross},
  title     = {Mutual Information Between Discrete and Continuous Data Sets},
  journal   = {PLOS ONE},
  year      = {2014},
  volume    = {9},
  number    = {2},
  pages     = {e87357}
}

@book{huber1981robust,
  author    = {Peter J. Huber},
  title     = {Robust Statistics},
  publisher = {Wiley},
  year      = {1981}
}

@article{hampel1971robust,
  author    = {Frank R. Hampel},
  title     = {A General Qualitative Definition of Robustness},
  journal   = {Annals of Mathematical Statistics},
  year      = {1971},
  volume    = {42},
  number    = {6},
  pages     = {1887--1896}
}

@book{molnar2020interpretable,
  author    = {Christoph Molnar},
  title     = {Interpretable Machine Learning},
  publisher = {Lulu.com},
  year      = {2020},
  note      = {Online book, accessed 2025}
}

@inproceedings{alvarezmelis2018robustness,
  author    = {David Alvarez{-}Melis and Tommi S. Jaakkola},
  title     = {On the Robustness of Interpretability Methods},
  booktitle = {Proc. ICML},
  year      = {2018},
  pages     = {242--250}
}

@inproceedings{yeh2019infidelity,
  author    = {Chih{-}Kuan Yeh and Cheng{-}Yu Hsieh and Arun P. Choudhury and Pradeep Ravikumar},
  title     = {On the (In)fidelity and Sensitivity of Explanations},
  booktitle = {Proc. NeurIPS},
  year      = {2019}
}

@inproceedings{greenwald2001quantiles,
  author    = {Michael Greenwald and Sanjeev Khanna},
  title     = {Space-Efficient Online Computation of Quantile Summaries},
  booktitle = {Proc. SIGMOD},
  year      = {2001},
  pages     = {58--66}
}

@misc{dunning2013tdigest,
  author    = {Ted Dunning and Otmar Ertl},
  title     = {Computing Extremely Accurate Quantiles Using t-Digests},
  howpublished = {arXiv preprint arXiv:1902.04023},
  year      = {2019}
}

@misc{lecun1998mnist,
  author    = {Yann LeCun and Corinna Cortes and Christopher J. C. Burges},
  title     = {The {MNIST} Database of Handwritten Digits},
  howpublished = {http://yann.lecun.com/exdb/mnist/},
  year      = {1998}
}

@misc{dua2017uci,
  author    = {Dheeru Dua and Casey Graff},
  title     = {{UCI} Machine Learning Repository},
  howpublished = {https://archive.ics.uci.edu/},
  year      = {2017}
}

@article{michalak2010correlation,
  title={Correlation based feature selection method},
  author={Michalak, Krzysztof and Kwasnicka, Halina},
  journal={International Journal of Bio-Inspired Computation},
  volume={2},
  number={5},
  pages={319--332},
  year={2010},
  publisher={Inderscience Publishers}
}

@misc{sklearnDatasets,
  author = {{scikit-learn developers}},
  title  = {scikit-learn Datasets (Iris, Wine, Diabetes, Digits, California Housing, 20 Newsgroups)},
  howpublished = {\url{https://scikit-learn.org/stable/datasets/}},
  year   = {2025},
  note   = {Accessed 2025-10-07}
}

@misc{openml,
  author = {Joaquin Vanschoren and others},
  title  = {OpenML: An open, collaborative platform for machine learning},
  howpublished = {\url{https://www.openml.org/}},
  year   = {2014},
  note   = {Accessed 2025-10-07}
}

@article{pedregosa2011scikit,
  author  = {Pedregosa, F. and Varoquaux, G. and Gramfort, A. and Michel, V. and Thirion, B. and Grisel, O. and Blondel, M. and Prettenhofer, P. and Weiss, R. and Dubourg, V. and Vanderplas, J. and Passos, A. and Cournapeau, D. and Brucher, M. and Perrot, M. and Duchesnay, E.},
  title   = {Scikit-learn: Machine Learning in {P}ython},
  journal = {Journal of Machine Learning Research},
  year    = {2011},
  volume  = {12},
  pages   = {2825--2830}
}

@misc{excirdemo2025,
  author       = {{ExCIR Authors}},
  title        = {Synthetic benchmark datasets: 
                  gene\_expression, sensor\_fusion, satellite\_ndvi, 
                  weather\_station, ecg\_heartbeat, eeg\_signals, audio\_mfcc, 
                  network\_topology, industrial\_process, social\_network, 
                  cyber\_security, stock\_technical, mixed\_modal, 
                  retail\_demand, traffic\_flow, energy\_consumption, 
                  synthetic\_clf, synthetic\_reg},
  year         = {2025},
  howpublished = {Included in the official ExCIR reproducibility code release},
  note         = {Deterministic synthetic controls and derived tabular/signal tasks used for scaling, stability, and agreement experiments.}
}

@article{jaccard1901etude,
  title={{\'E}tude comparative de la distribution florale dans une portion des Alpes et des Jura},
  author={Jaccard, Paul},
  journal={Bull Soc Vaudoise Sci Nat},
  volume={37},
  pages={547--579},
  year={1901}
}

@book{tukey1977eda,
  title     = {Exploratory Data Analysis},
  author    = {Tukey, John W.},
  year      = {1977},
  publisher = {Addison–Wesley},
  address   = {Reading, MA}
}

@book{wasserman2004,
  title   = {All of Statistics: A Concise Course in Statistical Inference},
  author  = {Wasserman, Larry},
  year    = {2004},
  publisher = {Springer}
}

@article{jaccard1912,
  title   = {The distribution of the flora in the alpine zone},
  author  = {Jaccard, Paul},
  journal = {New Phytologist},
  volume  = {11},
  number  = {2},
  pages   = {37--50},
  year    = {1912},
  doi     = {10.1111/j.1469-8137.1912.tb05611.x}
}

@article{spearman1904,
  title   = {The Proof and Measurement of Association Between Two Things},
  author  = {Spearman, Charles},
  journal = {The American Journal of Psychology},
  volume  = {15},
  number  = {1},
  pages   = {72--101},
  year    = {1904},
  doi     = {10.2307/1412159}
}

@article{kendall1938,
  title   = {A New Measure of Rank Correlation},
  author  = {Kendall, Maurice G.},
  journal = {Biometrika},
  volume  = {30},
  number  = {1/2},
  pages   = {81--93},
  year    = {1938},
  doi     = {10.1093/biomet/30.1-2.81}
}

@article{jeffreys1946,
  title   = {An Invariant Form for the Prior Probability in Estimation Problems},
  author  = {Jeffreys, Harold},
  journal = {Proceedings of the Royal Society of London. Series A},
  volume  = {186},
  number  = {1007},
  pages   = {453--461},
  year    = {1946},
  doi     = {10.1098/rspa.1946.0056}
}

@article{schonemann1966procrustes,
  title   = {A Generalized Solution of the Orthogonal Procrustes Problem},
  author  = {Sch{\"o}nemann, Peter H.},
  journal = {Psychometrika},
  volume  = {31},
  number  = {1},
  pages   = {1--10},
  year    = {1966},
  doi     = {10.1007/BF02289451}
}

@inproceedings{guo2017calibration,
  title={On Calibration of Modern Neural Networks},
  author={Guo, Chuan and Pleiss, Geoff and Sun, Yu and Weinberger, Kilian Q.},
  booktitle={ICML},
  year={2017}
}

@inproceedings{platt1999probabilistic,
  title={Probabilistic Outputs for Support Vector Machines and Comparisons to Regularized Likelihood Methods},
  author={Platt, John},
  booktitle={Advances in Large Margin Classifiers},
  year={1999}
}

@article{shannon1948,
  title={A Mathematical Theory of Communication},
  author={Shannon, Claude E.},
  journal={Bell System Technical Journal},
  volume={27},
  pages={379--423, 623--656},
  year={1948}
}

@article{kraskov2004,
  title={Estimating Mutual Information},
  author={Kraskov, Alexander and St{\"o}gbauer, Harald and Grassberger, Peter},
  journal={Physical Review E},
  volume={69},
  number={6},
  pages={066138},
  year={2004}
}

@article{ross2014,
  title={Mutual Information between Discrete and Continuous Data Sets},
  author={Ross, Brian C.},
  journal={PLoS ONE},
  volume={9},
  number={2},
  pages={e87357},
  year={2014}
}

\end{document}